\documentclass{article} 
\usepackage{iclr2027_conference,times}
\usepackage[T1]{fontenc}

\usepackage{amsmath,amsfonts,bm}

\def\eqref#1{equation~\ref{#1}}

\def\1{\bm{1}}

\DeclareMathAlphabet{\mathsfit}{\encodingdefault}{\sfdefault}{m}{sl}
\SetMathAlphabet{\mathsfit}{bold}{\encodingdefault}{\sfdefault}{bx}{n}

\usepackage{amsmath,listings,longtable,array,booktabs}
\usepackage{hyperref}
\usepackage{url}
\usepackage{amssymb}
\usepackage{array}
\usepackage{graphicx,booktabs,xcolor}
\title{FilmFlow: Agentic System for 3D Virtual Film Cinematography}

\author{%
\normalfont
Yiming Zhang$^{1,2}$\thanks{Equal contribution.} \quad
Jiarui Guo$^{2}$\footnotemark[1] \quad
Haojia Wei$^{2,3}$\footnotemark[1] \quad
Yifei Liu$^{4}$ \\
\normalfont
Yifan Yang$^{2}$ \quad
Hongjie Zhang$^{4}$ \quad
Xue Yang$^{2}$ \quad
Zhihang Zhong$^{2}$\thanks{Corresponding author.} \\[4pt]
\normalfont $^{1}$Cornell University \\
\normalfont $^{2}$Shanghai Jiao Tong University \\
\normalfont $^{3}$University of California, Los Angeles \\
\normalfont $^{4}$Shanghai Artificial Intelligence Laboratory
\\[6pt]
\href{https://visionary-laboratory.github.io/Filmflow/}{\texttt{https://visionary-laboratory.github.io/Filmflow/}}
}

\iclrfinalcopy
\begin{document}

\maketitle
\lhead{Preprint}
\vspace{-5mm}
\begin{figure}[h!]
\centering
\includegraphics[width=\linewidth]{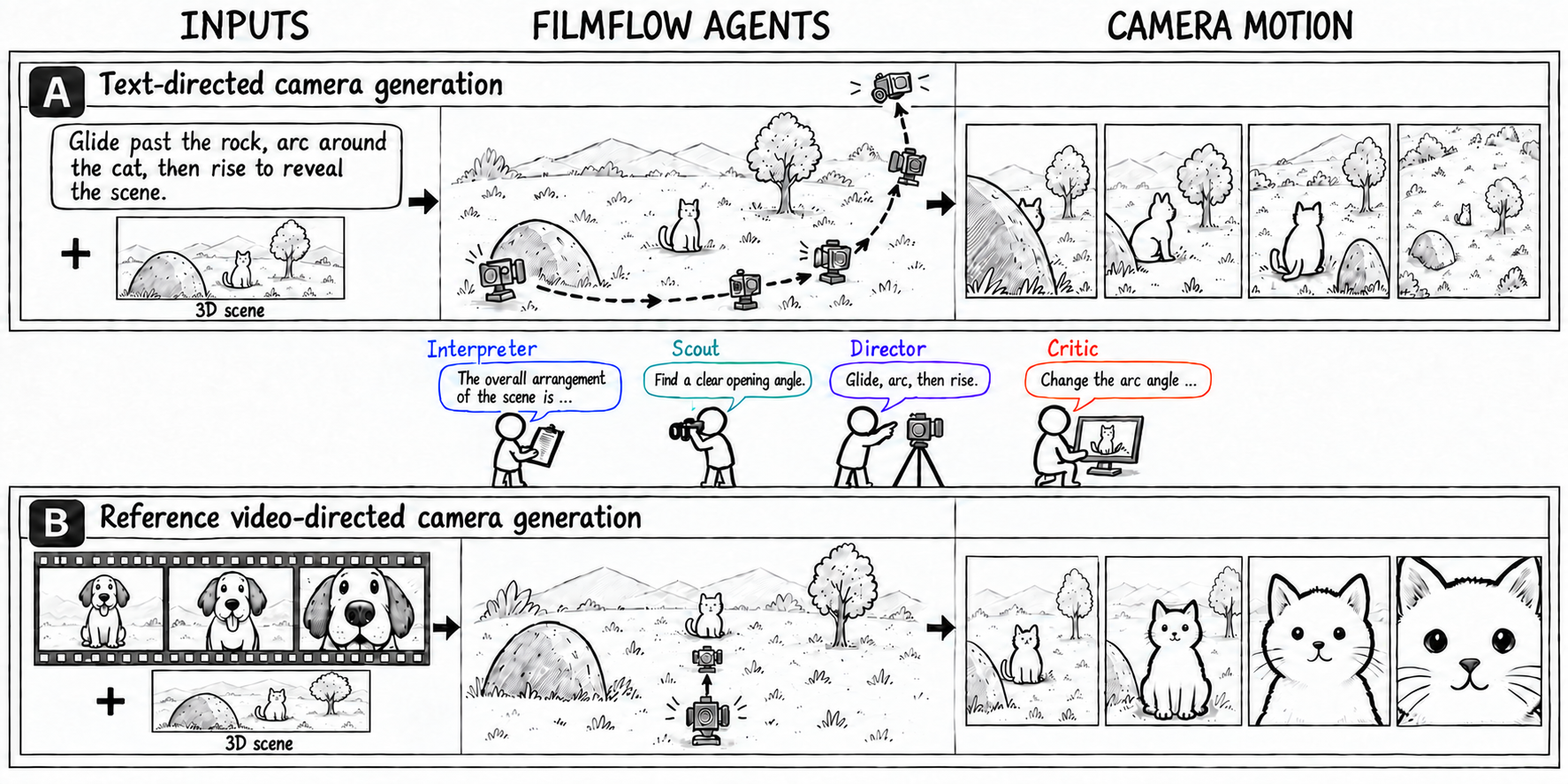}
\caption{\textbf{FilmFlow at a glance.}
Cooperating agents use visual feedback to iteratively refine
executable camera plans, enabling text-directed camera generation
and reference-video directed camera generation in static or dynamic 3D scenes.}
\label{fig:teaser}
\end{figure}
\vspace{-2mm}
\begin{abstract}
\vspace{-2mm}
Complex camera motion is central to visual storytelling, yet translating cinematographic intent into controllable, executable camera trajectories remains challenging. 
We introduce FilmFlow, an agentic framework that converts natural-language instructions or reference videos into explicit camera trajectories for dynamic 3D environments. 
To generate these trajectories, a Director agent constructs executable camera plans by composing high-level motions from concurrent atomic operations that jointly control translation, rotation, and framing. The Director is supported by three specialized agents: an Interpreter that extracts scene-grounded cinematographic intent, a Scout that selects initial viewpoints, and a Critic that evaluates rendered results and provides visual feedback for iterative plan refinement. 
We also introduce CinemaBench, comprising 150 Blender scenes from self-created and licensed sources, with five camera instructions per scene for 750 text-directed tasks and 60 reference videos associated with a subset of scenes. 
CinemaBench supports evaluation of both text-directed camera generation and video reference-directed camera generation. 
Across a diverse set of evaluated baselines, FilmFlow achieves the strongest benchmark performance and the highest user preference, demonstrating effective spatial reasoning, temporal coordination, and visual composition for executable 3D cinematography.
\end{abstract}


\vspace{-5mm}
\section{Introduction}
\vspace{-3mm}

Camera movement is a fundamental expressive device in filmmaking, shaping viewers’ attention, emotional engagement, and understanding of a scene \citep{brown2022cinematography}. A gradual push-in can emphasize a character’s reaction, while a pullback can reveal surrounding action and change the meaning of a moment. Achieving these narrative effects requires carefully coordinating camera movement, framing, and subject motion throughout a shot \citep{block2021visual}. Advances in spatial intelligence open up new possibilities for automating this coordination, enabling AI systems to reason about scene geometry and spatial relationships. Yet translating this understanding into cinematographically meaningful camera behavior remains challenging. Filmmakers often communicate their intentions through natural-language descriptions or reference footage, leaving the spatial and temporal decisions needed for execution implicit.

Prior work has made substantial progress toward automated cinematography, from rule-based and optimization-based camera control~\citep{christie2008camera} to learned trajectory generation conditioned on textual descriptions~\citep{jiang2024cinematographic}. More recently, ChatCam~\citep{liu2024chatcam} uses a language agent to interpret user requests and invoke camera-generation tools, while FilmAgent~\citep{xu2025filmagent} coordinates broader filmmaking workflows through multi-agent collaboration. CinemaTraj~\citep{li2026cinematraj} further grounds camera planning in 3D scene graphs, sequentially composing parametric motion primitives and optimizing their parameters to reduce collisions and occlusions in static scenes. 
Building on these advances, we address two coupled challenges: maintaining intended framing as subjects move, and coordinating concurrent camera operations with distinct timing. For example, a shot may orbit a moving subject while rising, then begin a push-in before the orbit ends. Such intent requires explicit control over how operations overlap and jointly determine camera behavior. The rendered result must also be assessed for composition and pacing, which geometric feasibility alone cannot ensure.

To address these challenges, we introduce FilmFlow, an agentic framework that generates executable camera trajectories in dynamic 3D scenes from natural-language instructions or reference videos. To generate these trajectories, a Director agent constructs structured camera plans by composing high-level motions from concurrent atomic operations. Three specialized agents support the Director: an Interpreter extracts scene-grounded cinematographic intent from the input, a Scout selects initial viewpoints, and a Critic evaluates executed plans and their rendered results. 
Because a poor opening constrains every later operation, the Scout
searches for this viewpoint before planning; and to reuse compositions
that have already succeeded, an in-context learning
mechanism~\citep{brown2020language} retrieves validated demonstrations
to guide the Director. 
We also introduce CinemaBench, comprising 150 Blender scenes from self-created and licensed sources, with five camera instructions per scene for 750 text-directed tasks and 60 reference videos associated with a subset of scenes. CinemaBench supports evaluation of both text-directed camera generation and video reference-directed camera generation. Across the evaluated baselines, FilmFlow achieves the strongest benchmark performance and the highest user preference.
Our contributions are threefold:
\vspace{-2mm}
\begin{itemize}
    \item \textbf{Agentic cinematography in dynamic scenes.}
    To our knowledge, FilmFlow is the first agentic framework to unify text-directed and video reference-directed camera generation in dynamic 3D scenes. The framework centers on a Director that constructs executable camera plans, assisted by specialized Interpreter, Scout, and Critic agents for scene-grounded intent extraction, initial viewpoint selection, and evaluation and refinement.

    \item \textbf{Concurrent motion representation and optimization.}
We introduce a structured camera-plan representation that composes high-level motions from concurrent atomic operations, with explicit parameters and temporal intervals to coordinate translation, rotation, and framing. A Critic-guided refinement loop revises individual operations, their
parameters, and their timing using execution and visual feedback, while
an in-context learning mechanism guides camera planning.

    \item \textbf{CinemaBench.}
    We introduce CinemaBench, comprising 150 Blender scenes, 750 text-directed tasks, and 60 reference videos associated with a subset of scenes, to evaluate both text-directed and video reference-directed camera generation. FilmFlow achieves the best performance across all reported automatic metrics and the highest overall preference in user studies for both text-directed and video reference-directed camera generation among the evaluated baselines.
\end{itemize}
\vspace{-4mm}
\section{Related Work}
\label{sec:related-work}
\vspace{-3mm}
\paragraph{Computational Cinematography and Camera Control.}
Computational cinematography translates narrative intent into camera control. Classical methods combine cinematic rules with geometric or image-space constraints~\citep{christie2008camera}, while Camera-on-rails incorporates character motion and framing preferences into smooth camera paths~\citep{galvane2015camera}. Learning-based approaches generate camera motion from text, including
diffusion-based synthesis with keyframing constraints~\citep{jiang2024cinematographic}
and character-aware trajectory generation trained on camera paths
extracted from films~\citep{courant2024et}. For video reference-directed camera generation, JAWS optimizes camera parameters and timing through differentiable rendering~\citep{wang2023jaws}, and subsequent work introduces cinematography-grounded losses~\citep{wang2026improved}. Auteur uses a domain-specific language for human-centric framing and predicts camera keyframes from language and human motion~\citep{kizil2026auteur}. Camera-controlled video generation conditions diffusion models on
camera trajectories~\citep{he2025cameractrl,wang2024motionctrl,bai2025recammaster},
which FilmFlow's explicit trajectories can directly supply. FilmFlow unifies text-directed generation and video reference-directed camera generation through executable camera planning and refinement in dynamic 3D scenes.
\vspace{-4mm}
\paragraph{Compositional Camera Motion and Trajectory Optimization.}
Structured representations define interpretable camera controls and optimization variables. Camera-on-rails optimizes movement along computed rails for smoothness and framing~\citep{galvane2015camera}. LAMP generates symbolic object and camera programs by combining cinematic primitives with spatial, temporal, and stylistic modifiers, supporting coordinated object--camera motion and combined controls within primitives~\citep{kizil2026lamp}. CinemaTraj grounds parametric primitives in a 3D scene graph, concatenates atomic segments sequentially, and optimizes trajectory parameters to reduce collisions and occlusions while preserving shot types~\citep{li2026cinematraj}. FilmFlow represents camera plans as independently parameterized atomic operations with explicitly overlapping temporal intervals. Execution checks and rendered feedback guide revisions to their parameters and timing, enabling coordinated actions such as orbiting, rising, and tracking a moving subject.
\vspace{-4mm}
\paragraph{Language Agents for Filmmaking and Visual Feedback.}
Language agents integrate reasoning, tool use, and environmental feedback. ReAct interleaves reasoning with actions and observations~\citep{yao2023react}, Reflexion uses verbal feedback to guide subsequent attempts~\citep{shinn2023reflexion}, and in-context learning specifies behavior through demonstrations~\citep{brown2020language}. In filmmaking, ChatCam connects conversational requests to trajectory-generation and anchor-selection tools~\citep{liu2024chatcam}, while FilmAgent coordinates a virtual film crew through role assignment and iterative script verification~\citep{xu2025filmagent}. PhotoFlow~\citep{guo2026photoflow} uses language agents for virtual photography, selecting single camera views in 3D scenes. Extending these mechanisms to continuous camera control requires linking visual deficiencies to actionable plan revisions. FilmFlow assigns intent interpretation, opening-view selection, and camera planning to the Interpreter, Scout, and Director, respectively. A shared Critic evaluates opening views and executed motion, combining stage-specific visual assessment with programmatic checks during trajectory evaluation. Its feedback guides viewpoint adjustments and operation-level revisions, making the executable camera plan the shared representation for planning, evaluation, and refinement.

\vspace{-3mm}

\section{Methods}
\vspace{-3mm}

Given a cinematographic input $x$, either a natural-language instruction
or a reference video, and a target 3D scene $S$, FilmFlow generates
executable camera trajectories through four specialized agents:
Interpreter, Scout, Director, and Critic. We describe the formulation
for a single continuous shot.
The Interpreter stage grounds the input in the target scene and
constructs a structured object description $\mathcal{G}$ and a
cinematographic script $\mathcal{B}$:
\begin{equation}
(\mathcal{G}, \mathcal{B}) =
A_{\mathrm{Interpreter}}(x,S).
\end{equation}
Here, $\mathcal{G}$ describes relevant scene objects and their
relationships, while $\mathcal{B}$ specifies the intended camera
behavior, duration, and target subjects. For video inputs, this stage
additionally analyzes the reference's motion and framing, and attaches
sampled reference frames and relative camera poses estimated by
LingBot-Map~\citep{chen2026geometric} as auxiliary evidence. These poses provide motion cues
with unverified metric scale; camera placement and movement are
resolved in the target scene.
The Scout proposes an initial camera state
$c^{(0)}=A_{\mathrm{Scout}}(\mathcal{G},\mathcal{B})$.
At scouting iteration $k$, the rendering operator $R$ produces opening images, which the Critic evaluates against $\mathcal{B}$:
\begin{equation}
(q_{\mathrm{view}}^{(k)},\mathcal{F}_{\mathrm{view}}^{(k)})
=
A_{\mathrm{Critic}}
\bigl(\mathcal{B},R(S,c^{(k)})\bigr),
\qquad
c^{(k+1)}
=
A_{\mathrm{Scout}}
\bigl(\mathcal{G},\mathcal{B},c^{(k)},
\mathcal{F}_{\mathrm{view}}^{(k)}\bigr).
\end{equation}
The search ends upon acceptance or exhaustion of its budget,
yielding the selected opening $c^*$.
Conditioned on this opening, the Director constructs an initial
camera plan:
\begin{equation}
\mathcal{P}^{(0)}
=
A_{\mathrm{Director}}(\mathcal{G},\mathcal{B},c^*).
\end{equation}
The plan specifies atomic operations and their temporal intervals,
which are composed into an executable trajectory and rendered.
At planning iteration $j$, the Critic assesses the resulting motion,
and its feedback guides plan revision:
\begin{equation}
(q_{\mathrm{motion}}^{(j)},\mathcal{F}_{\mathrm{motion}}^{(j)})
=
A_{\mathrm{Critic}}
\bigl(\mathcal{B},R(S,\mathcal{P}^{(j)})\bigr),
\qquad
\mathcal{P}^{(j+1)}
=
A_{\mathrm{Director}}
\bigl(\mathcal{G},\mathcal{B},c^*,\mathcal{P}^{(j)},
\mathcal{F}_{\mathrm{motion}}^{(j)}\bigr).
\end{equation}
For text inputs, the Critic assesses instruction fulfillment;
for video inputs, it compares timestamped reference and rendered
frames to assess camera-motion and framing fidelity. The Director
retains access to the reference evidence throughout refinement.
Both settings share the atomic-operation representation and
executor, while using task-specific evaluation criteria.
The selected opening remains fixed during motion refinement,
which terminates upon acceptance or exhaustion of the planning
budget. Here, $q$ denotes evaluation scores and $\mathcal{F}$
denotes revision feedback.

\begin{figure}[t!]
\centering
\includegraphics[width=\linewidth]{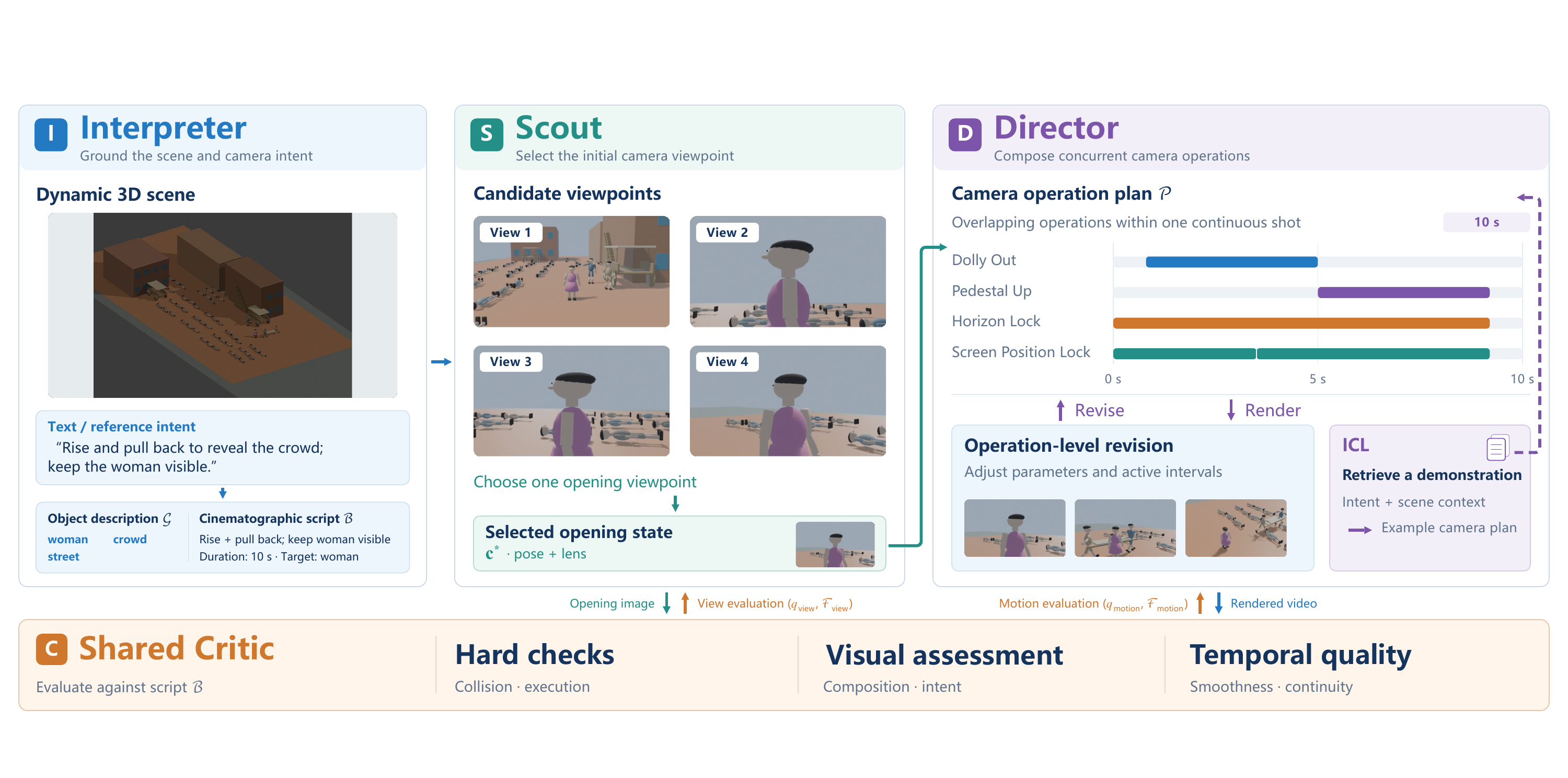}
\vspace{-10mm}
\caption{\textbf{System design of FilmFlow.} The Interpreter grounds
the instruction in the 3D scene; the Scout searches for an
opening viewpoint; the Director composes concurrent atomic operations
into a camera plan, optionally conditioned on a retrieved demonstration
(ICL). A shared Critic combines hard checks, visual assessment, and
temporal quality to guide both the Scout and Director.}
\label{fig:design_graph}
\end{figure}

\vspace{-3mm}
\subsection{Concurrent Atomic Camera Operations}
\label{sec:concurrent}
\vspace{-1.5mm}
\paragraph{Atomic camera operations.}
Complex cinematographic movements require coordinated changes in
camera position, orientation, and framing.
FilmFlow expresses such behavior through 22 atomic operations,
summarized in Table~\ref{tab:atomic_operations}.
For each shot, the Director constructs an operation timeline:
\begin{equation}
    \mathcal{P}=\{o_i\}_{i=1}^{N},
    \qquad
    o_i=(m_i,\boldsymbol{\theta}_i,[a_i,b_i]),
\end{equation}
where $m_i$ selects an operation,
$\boldsymbol{\theta}_i$ specifies its parameters and target references,
and $[a_i,b_i]$ defines its active interval.
Overlapping intervals allow multiple operations to jointly control
the camera.
For example, Orbit, Dolly In, and Pedestal Up can produce a movement
that circles a subject while approaching and ascending.
This decomposition preserves the meaning of each constituent action
and exposes its parameters and timing for subsequent refinement.

\begin{table}[t]
    \centering
    \small
    \vspace{-2mm}
    \caption{The 22 atomic camera operations supported by FilmFlow.}
    \label{tab:atomic_operations}
    \begin{tabular}{@{}p{0.20\linewidth}p{0.74\linewidth}@{}}
        \hline
        \textbf{Function} & \textbf{Operations} \\
        \hline
        Position &
        Dolly In, Dolly Out, Truck, Pedestal Up, Pedestal Down,
        Orbit, Path Follow, Position Follow, Height Follow,
        Distance Follow \\
        Orientation &
        Pan, Tilt, Aim Follow, Screen Position Lock,
        Roll, Horizon Lock \\
        Framing &
        Zoom, Frame Subject \\
        Focus &
        Focus Pull, Focus Follow \\
        Perturbation &
        Shake / Handheld \\
        Hold &
        Static Shot \\
        \hline
    \end{tabular}
    \vspace{-5mm}
\end{table}

\paragraph{Concurrent trajectory composition.}
We represent the camera state as
$\mathbf{c}(t)=(\mathbf{p}(t),\mathbf{R}(t),f(t))$,
where $\mathbf{p}(t)$ is its position, $\mathbf{R}(t)$ its orientation,
and $f(t)$ its focal length; additional lens parameters are omitted
for clarity.
The executor first expands each atomic operation into an absolute
camera-state sequence $\hat{\mathbf{c}}_i(t)$, using its parameters,
incoming camera state, and operation-specific interpolation.
It then extracts the relative change from the corresponding
reference state:
\begin{equation}
    \Delta\mathbf{c}_i(t)
    =
    \hat{\mathbf{c}}_i(t)
    \ominus
    \hat{\mathbf{c}}_i(t_i^{\mathrm{ref}}),
\end{equation}
where $\ominus$ denotes parameter-specific relative-change extraction
and $t_i^{\mathrm{ref}}$ is the operation start or the applicable
composition boundary.
The relative changes of compatible operations are combined at
each frame:
\begin{equation}
    \mathbf{c}(t)
    =
    \mathbf{c}_{\mathrm{base}}(t)
    \oplus
    \Delta\mathbf{c}_{i_1}(t)
    \oplus\cdots\oplus
    \Delta\mathbf{c}_{i_M}(t),
\end{equation}
where $\mathbf{c}_{\mathrm{base}}(t)$ is the reference or
tracking-derived camera state, and $\oplus$ composes relative
changes according to their parameter types and prescribed order.
This allows concurrent operations to jointly modify camera position,
orientation, and focal length.
The parameter-specific definitions of $\ominus$ and $\oplus$ are
provided in the appendix.
The resulting camera states are sampled per frame and baked into
animation keyframes.
\vspace{-3mm}
\subsection{Critic-Guided Evaluation and Refinement}
\label{sec:critic}
\vspace{-2mm}
\paragraph{Cinematographic evaluation and iterative refinement.}
The shared Critic evaluates rendered evidence against the script,
guiding the Scout's opening-view search and the Director's
operation-plan refinement.
Both stages use an equally weighted average of criterion-specific
scores in $[0,1]$:
\begin{equation}
q_\alpha(\mathcal{B},I)=\frac{1}{D_\alpha}\sum_{d=1}^{D_\alpha}q_{\alpha,d}(\mathcal{B},I),
\quad \alpha\in\{\mathrm{view},\mathrm{motion}\}.
\end{equation}
For scouting, $I=\mathcal{R}(\mathcal{S},\mathbf{c})$ and $D_{\mathrm{view}}=3$
(\textit{instruction alignment}, \textit{subject readability}, \textit{composition});
$\mathcal{F}_{\mathrm{view}}$ suggests adjustments to position $\mathbf{p}$, orientation $\mathbf{R}$, and
focal length $f$; the Scout perturbs the top-scoring candidates accordingly and fills the
remaining slots with newly sampled viewpoints.
For directing, $I=\mathcal{R}(\mathcal{S},\mathcal{P})$ and $D_{\mathrm{motion}}=6$,
adding \textit{spatial storytelling}, \textit{motion coherence}, and
\textit{cinematic quality}. Each executed plan first passes hard checks for
camera--scene collision and execution validity; $\mathcal{F}_{\mathrm{motion}}$ combines
any violations with operation-level diagnoses, which the Director uses to revise
$m_i$, $\boldsymbol{\theta}_i$, and $[a_i,b_i]$ with $\mathbf{c}^{*}$ fixed.
Scores are judged relative to the script, so intentional occlusion or delayed
revelation is not penalized. A plan is accepted only when its score exceeds the
threshold and the Critic approves it.

\vspace{-3mm}
\subsection{In-Context Cinematographic Planning}
\label{sec:icl}
\vspace{-2mm}
\paragraph{Experience collection and indexing.}
As an additional means of improving camera planning, FilmFlow equips
the Director with in-context demonstrations from previously validated
executions.
A shot is admitted to the example bank only if it passes technical
and execution checks, receives Critic acceptance, and satisfies
both overall and per-dimension quality thresholds.
Each stored example contains its task context, executable plan,
and evaluation outcome:
\begin{equation}
    e_j=(K_j,\mathcal{H}_j,\mathcal{P}_j,\mathcal{O}_j),
\end{equation}
where $\mathcal{H}_j$ includes the instruction, scene information,
and selected opening, and $\mathcal{O}_j$ records evaluation evidence
and feedback.
The retrieval key $K_j$ contains up to three keywords assigned by
the Interpreter from a shared vocabulary of visual goals, motion
requirements, and framing requirements.
These keywords describe the requested cinematography rather than
scene-specific object identities, allowing examples to be retrieved
across different scenes. The example bank is built only from development scenes, which are
disjoint from the test scenes used for evaluation in Sec.~\ref{sec:experiments}.

\paragraph{Example retrieval and conditioning.}
For a new script $\mathcal{B}$, the Interpreter produces
a keyword set $K$.
After opening-view selection, FilmFlow ranks eligible examples
by their Jaccard similarity to the query set:
\begin{equation}
    s(e_j,\mathcal{B})
    =\frac{|K_j\cap K|}{|K_j\cup K|},
\end{equation}
with an empty union contributing zero.
The highest-ranked valid example with positive similarity is
selected, using its Critic score to break ties.
Its compact representation $e^{*}$ contains the input conditions,
operation plan, and historical feedback, and augments the
Director's conditioning context:
\begin{equation}
\begin{aligned}
\widetilde{\mathcal{B}}
&=\operatorname{Concat}(\mathcal{B},e^{*}),\quad
\mathcal{P}^{(0)}=\mathcal{A}_{\mathrm{Director}}
\bigl(\mathcal{G},\widetilde{\mathcal{B}},\mathbf{c}^{*}\bigr).
\end{aligned}
\label{eq:icl_conditioning}
\end{equation}
Here, $\operatorname{Concat}$ combines the script and demonstration
in the input context without modifying the script.
The Director adapts the demonstrated operation choices and
temporal composition to the current scene and selected opening.
The example remains fixed throughout refinement, while current
constraints and Critic feedback take precedence.
If no qualifying example is found, planning proceeds without ICL,
with $\widetilde{\mathcal{B}}=\mathcal{B}$.

\providecommand{\expTBD}{\textcolor{gray}{\texttt{[TBD]}}}
\providecommand{\expNA}{\textemdash}
\providecommand{\expFrame}[3]{%
  \fbox{\begin{minipage}[c][#2][c]{\dimexpr#1-2\fboxsep-2\fboxrule\relax}
  \centering\scriptsize\color{gray}#3\end{minipage}}}


\newcommand{\ffplaceholder}[2]{%
  \fbox{%
    \parbox[c][#1][c]{%
      \dimexpr\linewidth-2\fboxsep-2\fboxrule\relax
    }{%
      \centering\small #2
    }%
  }%
}

\vspace{-3.5mm}
\section{Experiments}
\label{sec:experiments}
\vspace{-2mm}
\paragraph{CinemaBench.}
\vspace{-1mm}
CinemaBench comprises 150 Blender scenes (73 dynamic, 77 static), each
paired with five camera instructions of graded difficulty, and 60 reference videos for V2V
evaluation. We split it by scene into 50 development scenes, used only to build the ICL
example bank, and 100 test scenes (500 T2V and 40 V2V tasks), on which all results are
reported. Construction details are given in Appendix~\ref{sec:supp_benchmark}.
\vspace{-4mm}
\paragraph{Evaluation metrics.}
To avoid rewarding agreement with our own Critic, all outputs are scored exclusively by
external, off-the-shelf evaluators that share no weights, prompts, or training data with
FilmFlow; the same protocol is applied unchanged to every method and ablation. We report
means over all 500 T2V and 40 V2V test tasks. \textbf{Text Alignment} and \textbf{Motion
Quality} are computed with VideoReward~\citep{videoreward}, a reward model trained on
large-scale human preference judgments of video--prompt consistency and motion; overall
visual quality is measured by the composite \textbf{DOVER} score~\citep{dover}, which
jointly assesses aesthetic and technical quality, scaled to $[0,100]$. To target camera
behavior specifically, \textbf{Probe} applies the CameraBench camera-motion
model~\citep{camerabench}, trained on expert-annotated camera primitives, and reports the
VQAScore probability that a video matches its T2V instruction or its manually written V2V
motion description. We treat Probe as a complementary camera-motion signal and draw
conclusions only from trends that agree with the primary metrics and the user study
(Sec.~4.3). Higher is better for all metrics. All FilmFlow agents and the zero-shot VLM
baseline use gpt-6-astra with medium reasoning effort.

\vspace{-2.5mm}
\subsection{Comparison with Baselines}
\label{sec:baseline_comparison}
\vspace{-2mm}
\begin{table*}[h!]
\centering
\caption{\textbf{Quantitative comparison on CinemaBench.}
Results are averaged over all 500 T2V and 40 V2V tasks.
Align and Motion report VideoReward text alignment and motion
quality, respectively; Quality reports the DOVER composite
score (0--100).
Higher is better for all metrics.
Bold marks the best score in each column.}
\label{tab:baseline_comparison}
\small
\setlength{\tabcolsep}{2pt}
\begin{tabular*}{\textwidth}{
@{\extracolsep{\fill}}
lrrrr@{\hspace{10pt}}lrrrr@{}
}
\toprule
\multicolumn{5}{c}{\textit{T2V}}
& \multicolumn{5}{c}{\textit{V2V}} \\
\cmidrule(r{5pt}){1-5}
\cmidrule(l{5pt}){6-10}
Method
& Align. $\uparrow$
& Quality $\uparrow$
& Motion $\uparrow$
& Probe $\uparrow$
& Method
& Align. $\uparrow$
& Quality $\uparrow$
& Motion $\uparrow$
& Probe $\uparrow$ \\
\midrule
LAMP
& -1.747 & 41.54 & 0.209 & 0.564
& Zero-shot VLM
& -1.417 & 35.86 & 0.041 & 0.502 \\
GenDoP
& -1.768 & 39.63 & 0.257 & 0.597
& LingBot-Map
& -1.472 & 35.55 & 0.278 & 0.521 \\
CinemaTraj
& -1.788 & 44.65 & -0.068 & 0.618
& VGGT-$\Omega$
& -1.491 & 34.88 & 0.176 & 0.533 \\
\textbf{FilmFlow}
& \textbf{-1.461}
& \textbf{47.60}
& \textbf{0.555}
& \textbf{0.829}
& \textbf{FilmFlow}
& \textbf{-1.224} & \textbf{41.77} & \textbf{0.289} & \textbf{0.750} \\
\bottomrule
\end{tabular*}
\end{table*}

\begin{figure*}[h!]
\centering
\includegraphics[width=\textwidth]{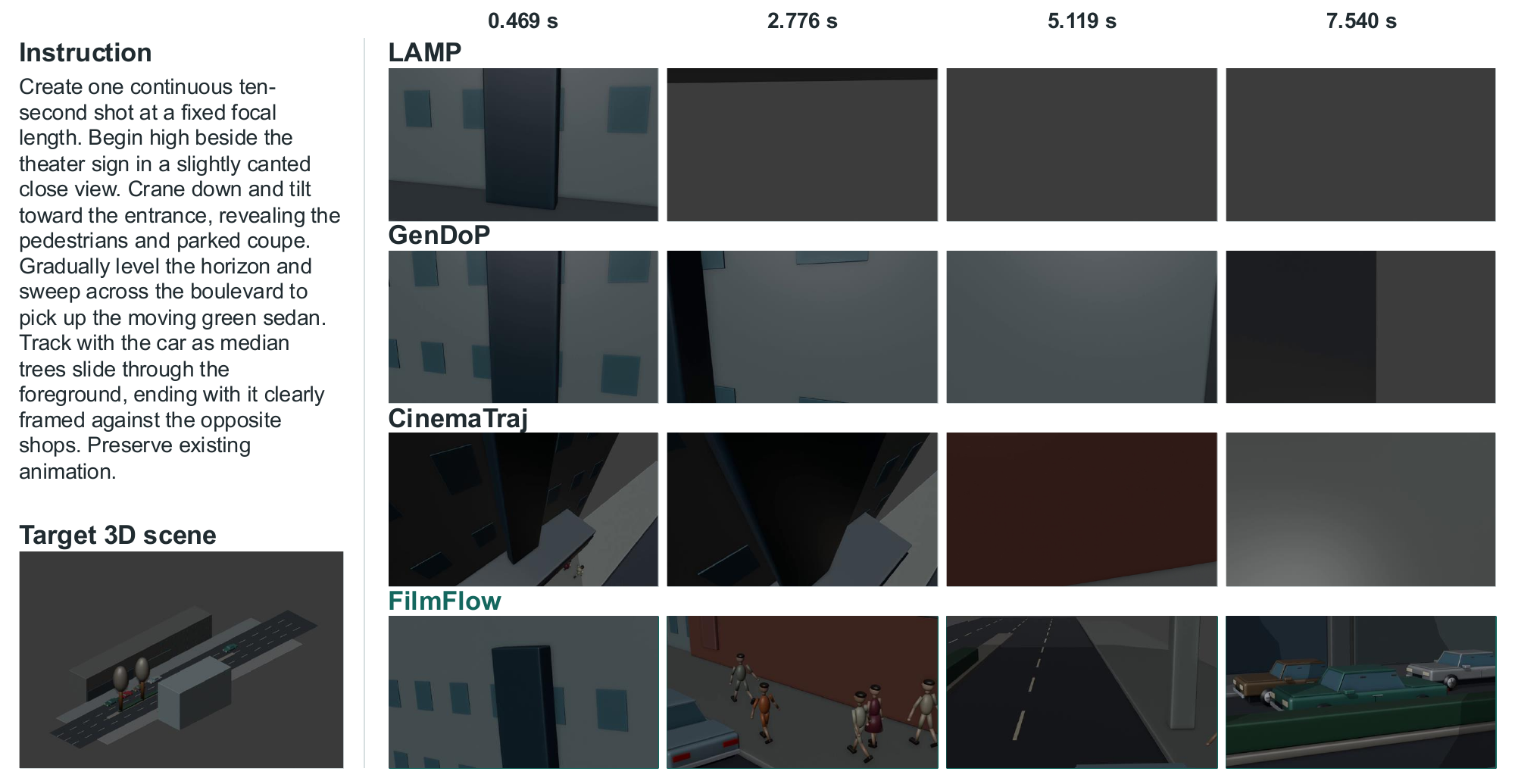}
\caption{\textbf{Qualitative comparison for text-directed camera generation.}
The instruction and target 3D scene are shown on the left;
rows show each method's output at the indicated timestamps.
While the baselines mostly show walls or obstructed views,
FilmFlow reveals the pedestrians, turns toward the boulevard,
and brings the target green sedan into view.}
\vspace{-5.5mm}
\label{fig:t2v_comparison}
\end{figure*}

\begin{figure*}[h!]
\centering
\includegraphics[width=\textwidth]{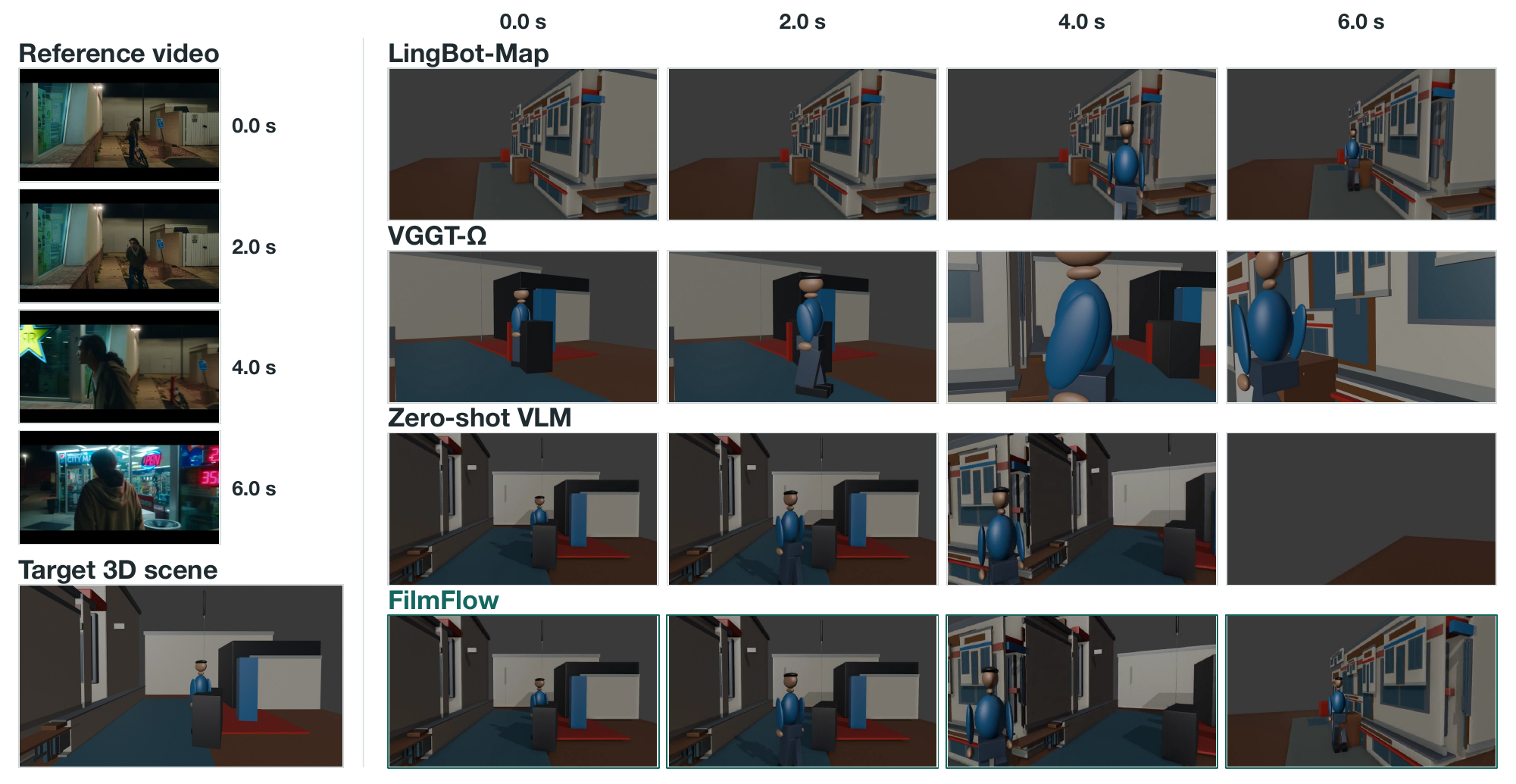}
\vspace{-5mm}
\caption{\textbf{Qualitative comparison for video reference-directed camera generation.}
Reference frames and the target 3D scene are shown on the left;
rows on the right show each method's output at the indicated
timestamps.
FilmFlow adjusts the viewpoint as the subject moves and keeps
the subject visible across the sampled frames.
LingBot-Map misses the subject in the opening frames,
VGGT-$\Omega$ crops the subject in later frames,
and the zero-shot VLM loses the subject in the final frame.
The comparison concerns camera behavior rather than scene appearance.}
\vspace{-6mm}
\label{fig:v2v_comparison}
\end{figure*}

For T2V, we compare FilmFlow with LAMP~\citep{kizil2026lamp},
GenDoP~\citep{zhang2025gendop}, and CinemaTraj~\citep{li2026cinematraj}
on CinemaBench, with all methods receiving the same task instruction
and target scene.
For V2V, we compare against direct camera-pose extraction using
LingBot-Map~\citep{chen2026geometric} and
VGGT-$\Omega$~\citep{wang2026vggtomega},
as well as zero-shot camera planning with a VLM,
using the same reference video and target scene.
For the pose-extraction baselines, the estimated camera trajectories
are converted to the target scene's coordinate convention and
applied for rendering. Full adaptation details for all baselines are given in Appendix~B.3.
Table~\ref{tab:baseline_comparison} reports the two tasks in separate
blocks using the same four metrics: VideoReward text alignment, DOVER
overall quality, VideoReward motion quality, and the auxiliary Probe score.
Alignment measures agreement with the text instruction for T2V
and the reference motion description for V2V, and is compared
only within each task block.
Figures~\ref{fig:t2v_comparison} and~\ref{fig:v2v_comparison}
show qualitative results at matching normalized timestamps,
illustrating the progression of camera actions and composition.
The text instruction or reference clip is displayed separately
from the outputs, and full videos accompany the visualizations.

\vspace{-2mm}
\subsection{Ablation Study}
\label{sec:ablation}
\vspace{-2mm}

\begin{table*}[h!]
\centering
\caption{\textbf{Ablation study on CinemaBench.}
Full FilmFlow achieves the highest alignment, video quality,
and motion quality scores in both tasks, and the highest
T2V Probe score.
Probe is the mean match probability from the CameraBench-based
evaluator.
Higher is better; bold marks the best score in each column.}
\label{tab:ablation}
\small
\setlength{\tabcolsep}{2pt}
\begin{tabular*}{\textwidth}{
@{\extracolsep{\fill}}
lrrrr@{\hspace{10pt}}rrrr@{}
}
\toprule
& \multicolumn{4}{c}{\textit{T2V}}
& \multicolumn{4}{c}{\textit{V2V}} \\
\cmidrule(r{5pt}){2-5}
\cmidrule(l{5pt}){6-9}
Variant
& Align. $\uparrow$
& Quality $\uparrow$
& Motion $\uparrow$
& Probe $\uparrow$
& Align. $\uparrow$
& Quality $\uparrow$
& Motion $\uparrow$
& Probe $\uparrow$ \\
\midrule
w/o Scout
& -1.542 & 46.11 & 0.387 & 0.770
& -1.344 & 40.77 & 0.243 & 0.673 \\

Sequential-only Operations
& -1.547 & 45.06 & 0.536 & 0.775
& -1.314 & 40.78 & 0.278 & 0.717 \\

w/o Critic
& -1.499 & 45.25 & 0.430 & 0.770
& -1.288 & 37.21 & 0.191 & 0.716 \\

w/o ICL
& -1.474 & 46.12 & 0.447 & 0.724
& -1.296 & 39.04 & 0.172 & \textbf{0.766} \\

\textbf{Full FilmFlow}
& \textbf{-1.461}
& \textbf{47.60}
& \textbf{0.555}
& \textbf{0.829}
& \textbf{-1.224}
& \textbf{41.77}
& \textbf{0.289}
& 0.750 \\
\bottomrule
\end{tabular*}
\end{table*}

\vspace{-2mm}
\begin{figure*}[h!]
\centering
\includegraphics[width=\textwidth]{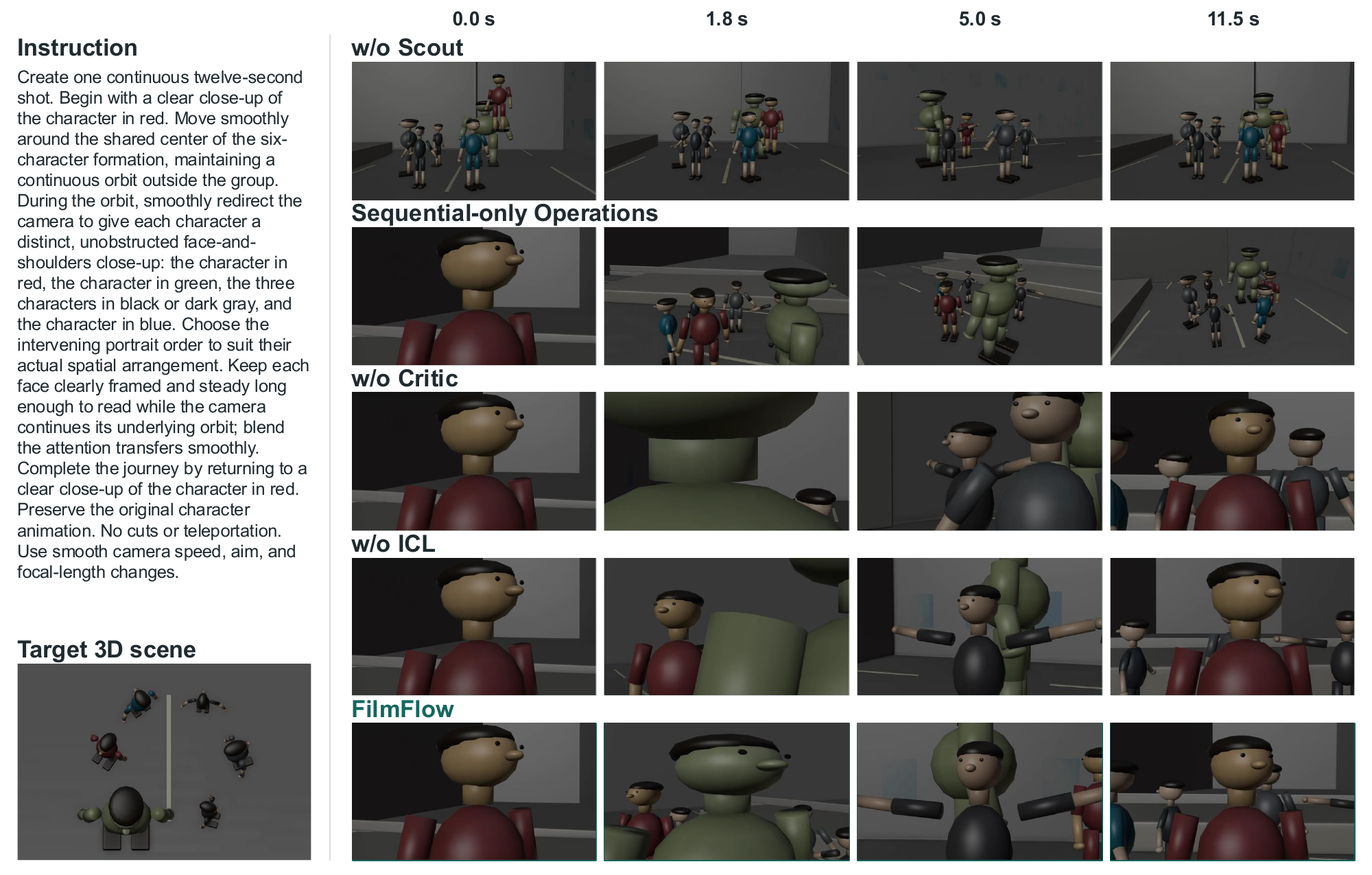}
\caption{\textbf{Qualitative ablation for text-directed camera generation.}
FilmFlow shows close views of the red, green, and dark-clothed
characters before returning to the red character.
Without Scout, the opening remains a distant group view;
without Critic or ICL, foreground characters obstruct the
intended subject in intermediate frames.
The instruction and target scene are shown on the left;
columns show frames at the indicated timestamps.}
\label{fig:t2v_ablation}
\end{figure*}

\begin{figure*}[h!]
\centering
\includegraphics[width=\textwidth]{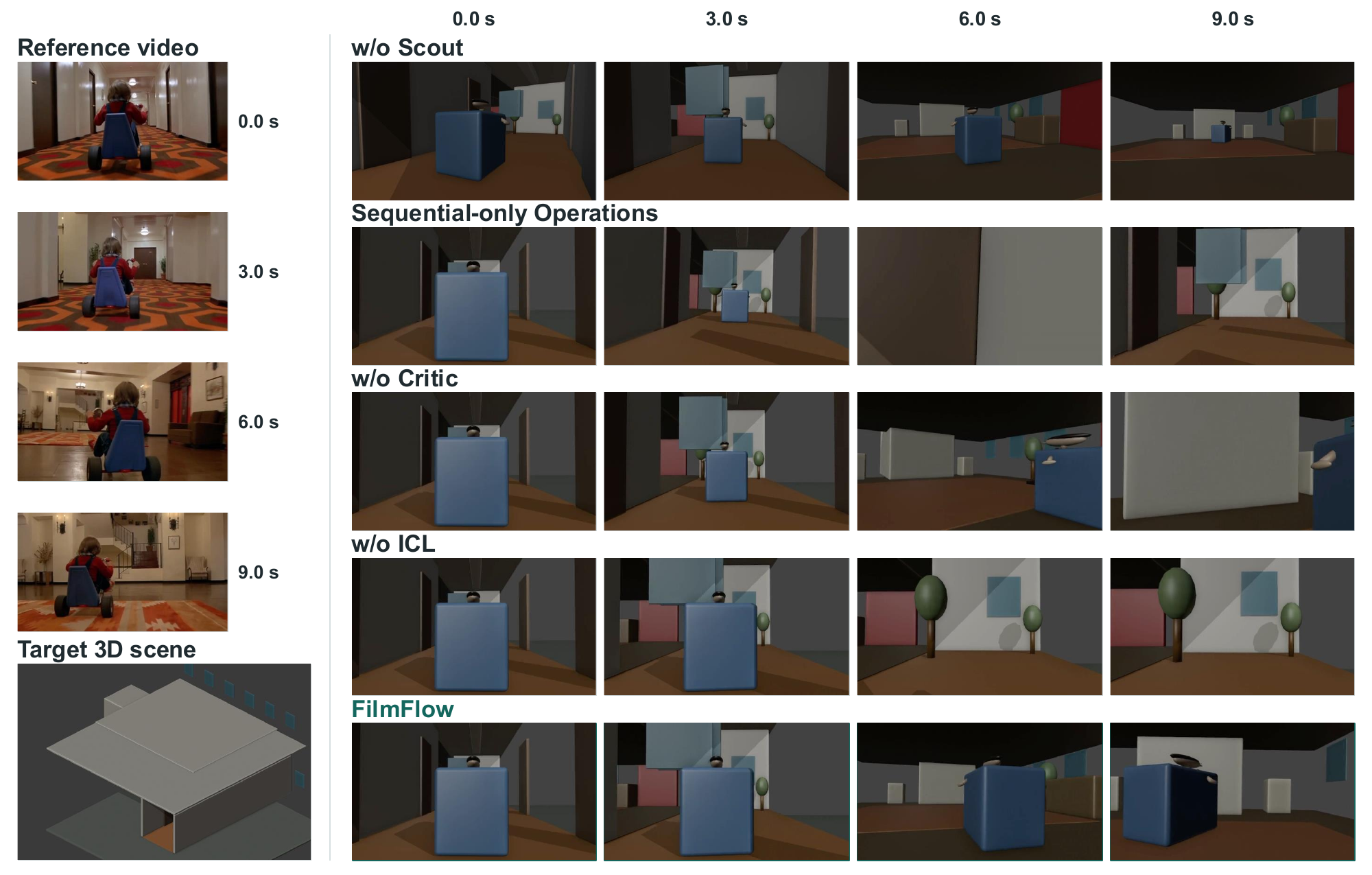}
\caption{\textbf{Qualitative ablation for video reference-directed camera generation.}
FilmFlow maintains more consistent target framing across the
sampled frames, while the ablated variants show less consistent
subject scale, obstructed views, or loss of the target.
Reference frames and the target 3D scene are shown on the left;
columns show outputs at the indicated timestamps.}
\label{fig:v2v_ablation}
\end{figure*}

We examine the contributions of opening-viewpoint selection,
concurrent atomic camera operations, iterative critique, and in-context
demonstrations.
The \textbf{w/o Scout} variant removes the separate opening search; the Director specifies the initial camera state as part of its plan.
The \textbf{Sequential-only Operations} variant retains the same atomic-operation library but prohibits overlapping execution intervals: multiple operations may be arranged sequentially within a shot, while at most one operation is active at any time.
The \textbf{W/o Critic} disables Critic-guided revision while
retaining programmatic validity checks;
and \textbf{w/o ICL} removes retrieved demonstrations from
the Director's context.
Table~\ref{tab:ablation} compares these variants with
\textbf{Full FilmFlow} on alignment, overall video quality,
motion quality, and the auxiliary Probe score,
with comparisons made within each task.
The external evaluation protocol remains fixed across variants,
so removing the internal Critic does not change how outputs
are scored.
Probe is an exploratory soft score, not a pass rate.
Full FilmFlow achieves the best score in every column except V2V
Probe, where w/o ICL is slightly higher (0.766 vs.\ 0.750); given
the auxiliary role of Probe and the consistent gains on all other
V2V metrics, we do not read this small gap as evidence against ICL.
Figures~\ref{fig:t2v_ablation} and~\ref{fig:v2v_ablation}
provide qualitative comparisons at matching timestamps
to assess opening composition, temporal coordination,
subject visibility, and agreement with the instruction
or reference motion.

\vspace{-2.5mm}
\subsection{User Study}
\label{sec:user_study}

\vspace{-2mm}
\label{sec:user_study}
We conduct blinded user studies for T2V and V2V separately, with
method identities hidden and presentation order randomized.
Forty-two participants were each assigned 10 T2V and 10 V2V trials,
randomly sampled from the T2V and V2V test tasks. Each trial shows the text instruction (T2V) or reference
video (V2V) alongside the outputs of all four methods. Participants
rate each output on a five-point scale for camera-behavior alignment,
overall visual quality (aesthetics and technical fidelity), and motion
smoothness; judge whether it realizes the specified camera behavior
(\emph{pass} or \emph{fail}); and rank the outputs by overall
preference. For V2V, alignment and pass/fail judgments concern the
reference camera behavior rather than scene appearance.
All statistics are computed over the 420 T2V and 397 V2V
completed rankings.
Table~\ref{tab:user_study_ratings} reports mean ratings, and
Table~\ref{tab:user_study} reports the human pass rate (the fraction
of judgments labeled \emph{pass}) and the Top-1 rate (the fraction of
rankings in which a method is placed first). 

\begin{table*}[h!]
\centering
\vspace{-2mm}
\caption{
\textbf{User study on CinemaBench.}
FilmFlow has the highest pass rate in both tasks and is ranked
first in 61.0\% of T2V judgments and 63.2\% of V2V judgments.
Pass Rate is the proportion of completed judgments marked
as pass; Top-1 Rate is the proportion of rankings in which
a method is placed first.
Higher is better; bold marks the best result within each task
and metric.}
\label{tab:user_study}
\small
\setlength{\tabcolsep}{8pt}
\begin{tabular*}{\textwidth}{@{}p{0.18\textwidth}*{2}{>{\centering\arraybackslash}p{\dimexpr(0.64\textwidth-20pt-8\tabcolsep)/4\relax}}@{\hspace{20pt}}p{0.18\textwidth}*{2}{>{\centering\arraybackslash}p{\dimexpr(0.64\textwidth-20pt-8\tabcolsep)/4\relax}}@{}}
\toprule
\multicolumn{3}{@{}c@{\hspace{20pt}}}{\textit{T2V}} & \multicolumn{3}{c}{\textit{V2V}} \\
\cmidrule(r{20pt}){1-3}\cmidrule(l{0pt}){4-6}
Method & \mbox{Pass Rate\,$\uparrow$} & \mbox{Top-1 Rate\,$\uparrow$}
& Method & \mbox{Pass Rate\,$\uparrow$} & \mbox{Top-1 Rate\,$\uparrow$} \\
\midrule
LAMP       & 42.1\% & 12.1\% & Zero-shot VLM & 63.0\% & 2.3\% \\
GenDoP     & 46.2\% & 5.7\%  & LingBot-Map   & 77.3\% & 17.4\% \\
CinemaTraj & 63.8\% & 21.2\% & VGGT-$\Omega$ & 73.3\% & 17.1\% \\
FilmFlow   & \textbf{84.3\%} & \textbf{61.0\%} & FilmFlow      & \textbf{82.1\%} & \textbf{63.2\%} \\
\bottomrule
\end{tabular*}
\end{table*}

\begin{table*}[h!]
\centering
\vspace{-3mm}
\caption{\textbf{Human ratings on CinemaBench.}
FilmFlow receives the highest mean ratings for alignment,
visual quality, and motion smoothness in both tasks.
Align., Quality, and Motion denote the corresponding ratings
on a five-point scale.
For V2V, alignment measures how well the camera behavior
matches the reference, rather than similarity in scene appearance.
Higher is better; bold marks the best result within each task
and criterion.}
\label{tab:user_study_ratings}
\small
\setlength{\tabcolsep}{3pt}
\begin{tabular*}{\textwidth}{
@{\extracolsep{\fill}}
lrrr@{\hspace{16pt}}lrrr@{}
}
\toprule
\multicolumn{4}{c}{\textit{T2V}}
& \multicolumn{4}{c}{\textit{V2V}} \\
\cmidrule(r{8pt}){1-4}
\cmidrule(l{8pt}){5-8}
Method
& Align. $\uparrow$
& Quality $\uparrow$
& Motion $\uparrow$
& Method
& Align. $\uparrow$
& Quality $\uparrow$
& Motion $\uparrow$ \\
\midrule
LAMP
& 2.6 & 2.9 & 2.7
& Zero-shot VLM
& 2.5 & 2.8 & 2.7 \\
GenDoP
& 2.8 & 3.0 & 2.9
& LingBot-Map
& 3.2 & 3.5 & 3.3 \\
CinemaTraj
& 3.5 & 3.7 & 3.6
& VGGT-$\Omega$
& 3.4 & 3.6 & 3.5 \\
\textbf{FilmFlow}
& \textbf{4.2} & \textbf{4.4} & \textbf{4.3}
& \textbf{FilmFlow}
& \textbf{4.0} & \textbf{4.3} & \textbf{4.0} \\
\bottomrule
\end{tabular*}
\end{table*}
\vspace{-4.5mm}
\section{Conclusion}
\vspace{-2mm}
We presented FilmFlow, an agentic framework that generates executable
camera trajectories in dynamic 3D scenes from natural-language
instructions or reference videos.
By composing concurrent atomic operations and refining plans through
execution and visual feedback, FilmFlow coordinates camera motion,
subject tracking, and framing to realize cinematographic intent.
In-context demonstrations further guide operation selection and
temporal composition.
We also introduced CinemaBench to evaluate camera generation under
both input modalities.
Experiments show that FilmFlow achieves the strongest benchmark
performance and highest user preference among the evaluated baselines.
Future work will investigate whether smaller models can achieve
comparable performance through task-specific fine-tuning,
and extend FilmFlow to jointly automate scene construction
and camera planning.
\newpage

\subsection*{AI Use Statement}

We used generative AI tools to improve the clarity and
readability of the manuscript, assist with debugging code,
generate synthetic datasets, and create the teaser figure.
The authors reviewed the AI-assisted text, code changes,
and teaser figure, and checked the generated data for
consistency with the intended experimental setup.
The authors take full responsibility for the final content
of this work, including the figures, datasets, implementation,
and reported results.

\subsection*{Ethics Statement}
CinemaBench includes third-party scene assets and reference videos.
Our release will preserve source attribution and respect applicable
licenses and redistribution restrictions. Materials without
redistribution permission will be excluded, with source links and
acquisition instructions provided where possible. The user study
received institutional ethics clearance. All participants provided
informed consent, and their responses were anonymized to protect
participant privacy.

\subsection*{Reproducibility Statement}
We will publicly release the FilmFlow implementation, CinemaBench annotations, along with instructions
for obtaining the required scene assets, to facilitate reproduction
of our results.

\bibliographystyle{iclr2027_conference}
\bibliography{iclr2027_conference}

\newpage
\appendix

\section{Technical Details of the Agentic System}
\label{sec:supp_technical}

\subsection{Agent Roles and Structured Interfaces}
\label{sec:supp_interfaces}

We illustrate the agent interfaces using a running T2V example, linking each variable in the method overview to a concrete prompt input, structured response, or rendered observation. Starting from the cinematographic input $x$ and scene observations extracted from $S$, we show the Interpreter's object description $\mathcal{G}$ and cinematographic script $\mathcal{B}$; the Scout's candidate camera states $c^{(k)}$, rendered opening views $R(S,c^{(k)})$, and selected opening $c^*$; and the Director's initial and revised plans $\mathcal{P}^{(j)}$ and $\mathcal{P}^{(j+1)}$. The Critic examples include both opening-view and motion evaluation, showing the evidence used to obtain scores $q_{\mathrm{view}}$ and $q_{\mathrm{motion}}$ and feedback $\mathcal{F}_{\mathrm{view}}$ and $\mathcal{F}_{\mathrm{motion}}$. A supplementary note below summarizes the reference-specific inputs and evaluation used for V2V.

The T2V example requests a dolly zoom toward a yellow-clothed figure on a highway. Prompt excerpts are abridged from the implementation, and response excerpts and numerical values are taken from the corresponding execution logs. Only relevant fields are shown, long identifiers are omitted where unnecessary, and numerical values are rounded for readability. These excerpts illustrate the interfaces rather than constitute complete executable requests. Visual inputs are described in the prompt excerpts and omitted from this presentation; camera states and rendered observations are produced by the search and execution modules.

\paragraph{Interpreter: from input to grounded intent}

The input $x$ is the following instruction. The scene $S$ is an existing Blender scene; the Interpreter receives its object inventory rather than the scene file as text.

\begin{lstlisting}[basicstyle=\ttfamily\footnotesize,breaklines=true,breakatwhitespace=false,columns=fullflexible,keepspaces=true,showstringspaces=false]
USER INSTRUCTION
One continuous ten-second shot. Center the yellow-clothed figure's upper body with the entire rounded hairstyle visible and the road divider receding on the left. Dolly forward while zooming out to hold the figure's apparent size as the roadway opens in perspective. Keep the divider clear of the torso and let its changing angle reveal depth. Ease the approach and zoom together, holding the final balanced portrait for one second.
\end{lstlisting}

Relevant observations identify \texttt{hero\_01} as the yellow-clothed figure, \texttt{median} as the road divider, and \texttt{freeway\_deck} as the roadway. For example, the measured bounding-box centers are $(2.1826,-4.2000,2.2473)$, $(0,55,0.54)$, and $(0,55,-0.169)$, respectively. The request specifies one continuous shot lasting 10 seconds. The Interpreter receives the following role instructions together with this scene inventory and $x$:

\begin{lstlisting}[basicstyle=\ttfamily\footnotesize,breaklines=true,breakatwhitespace=false,columns=fullflexible,keepspaces=true,showstringspaces=false]
SYSTEM EXCERPT
You are the Interpreter. Ground the scene and schedule shots,
but do not construct camera plans.
With shot_count=1 all requested actions belong to ONE continuous
shot; do not split it.
instruction describes cinematic intent; opening_intent describes
the initial visual composition, not an exact camera pose.
Do not output operations, camera coordinates, focal lengths, paths,
cut commands, shot IDs, start/end times or source time edits.
\end{lstlisting}

The following fields of the validated response instantiate $\mathcal{G}$ and $\mathcal{B}$. The shot's instruction field retains $x$ verbatim and is omitted below to avoid repetition; its 240-frame duration corresponds to 10 seconds at 24 fps.

\begin{lstlisting}[basicstyle=\ttfamily\footnotesize,breaklines=true,breakatwhitespace=false,columns=fullflexible,keepspaces=true,showstringspaces=false]
{
  "scene_understanding": {
    "subjects": ["hero_01", "median", "freeway_deck"],
    "opening_target": "hero_01",
    "final_target": "hero_01",
    "motion_objects": ["hero_01"],
    "virtual_objects": []
  },
  "shots": [
    {
      "opening_intent": "A centered upper-body portrait of hero_01, with the entire rounded hairstyle visible. The road divider median recedes on the left, clearly separated from the torso, with freeway_deck providing the roadway context.",
      "duration_frames": 240,
      "keywords": ["dolly_zoom_in", "perspective_change", "stable_subject_size"]
    }
  ]
}
\end{lstlisting}

Thus $\mathcal{G}$ identifies the figure, divider, and roadway and distinguishes the animated subject from static context; $\mathcal{B}$ specifies the requested dolly zoom, opening portrait, shot duration, and final hold. The Interpreter does not prescribe the eventual operation timeline.

\paragraph{Scout and opening Critic: candidate views and selection}

The Scout generates candidate camera states programmatically and submits their rendered images, target information, and geometric measurements to the opening Critic. To illustrate $c^{(k)}\rightarrow c^{(k+1)}$, we follow one candidate lineage within the multi-candidate search. Its initial state $c^{(0)}$ is:

\begin{lstlisting}[basicstyle=\ttfamily\footnotesize,breaklines=true,breakatwhitespace=false,columns=fullflexible,keepspaces=true,showstringspaces=false]
{
  "candidate_id": "opening_r00_013",
  "camera_position": [-2.1643, 0.1103, 3.2248],
  "look_at": [2.1826, -4.2, 2.7704],
  "lookup": [0, 0, 1],
  "focal_length_mm": 53.5887
}
\end{lstlisting}

Here, \texttt{look\_at} and \texttt{lookup} specify the viewing direction and up vector. The rendered opening $R(S,c^{(0)})$ is attached to the prompt with its candidate identifier; the full request includes the other candidates in the same round.

\begin{lstlisting}[basicstyle=\ttfamily\footnotesize,breaklines=true,breakatwhitespace=false,columns=fullflexible,keepspaces=true,showstringspaces=false]
PROMPT EXCERPT
Instruction: [the instruction and opening intent above]
Opening target assembly: hero_01. Evaluate the opening only.
Acceptance threshold: 0.7; selection uses score alone.

For each attached candidate image, return its candidate_id,
score, diagnosis, and adjustment suggestions.
\end{lstlisting}

The response below provides $q_{\mathrm{view}}^{(0)}=0.55$ and $\mathcal{F}_{\mathrm{view}}^{(0)}$ for this candidate:

\begin{lstlisting}[basicstyle=\ttfamily\footnotesize,breaklines=true,breakatwhitespace=false,columns=fullflexible,keepspaces=true,showstringspaces=false]
{
  "candidate_id": "opening_r00_013",
  "score": 0.55,
  "diagnosis": "The yellow figure and complete hairstyle are readable in a three-quarter view, but the crop includes the skirt and partial legs rather than emphasizing the upper body. Large headroom reduces portrait strength. The diagonal barrier crosses directly behind the torso instead of remaining clearly separated on the left.",
  "adjustment": {
    "azimuth_delta_deg": -25,
    "radius_scale": 0.75,
    "height_delta_scales": 0.1,
    "focal_scale": 1.2,
    "aim_height_delta_scales": -0.05
  }
}
\end{lstlisting}

The next round includes a feedback-guided perturbation of this candidate, with $c^{(1)}$ given below. The updated view receives $q_{\mathrm{view}}^{(1)}=0.68$.

\begin{lstlisting}[basicstyle=\ttfamily\footnotesize,breaklines=true,breakatwhitespace=false,columns=fullflexible,keepspaces=true,showstringspaces=false]
{
  "candidate_id": "opening_r01_003",
  "parent_id": "opening_r00_013",
  "camera_position": [2.3449, 0.124, 3.2714],
  "look_at": [2.1826, -4.2, 2.6887],
  "lookup": [0, 0, 1],
  "focal_length_mm": 60.2427,
  "score": 0.68
}
\end{lstlisting}

The Critic identifies improved facial readability and roadway perspective but still notes incorrect divider placement. No candidate reaches the 0.70 threshold within the two-round budget. The search therefore selects this highest-scoring view as the provisional opening $c^*=c^{(1)}$, rather than marking it as threshold-accepted. Its pose and focal length remain fixed during subsequent motion-plan refinement.

\paragraph{Director: constructing the initial operation plan}

The Director receives $\mathcal{B}$, measured scene context derived from $\mathcal{G}$, the selected opening $c^*$ and its RGB image, the atomic-operation codebook, and any retrieved demonstration. The initial request has no previous timeline or refinement feedback. Its role instructions include:

\begin{lstlisting}[basicstyle=\ttfamily\footnotesize,breaklines=true,breakatwhitespace=false,columns=fullflexible,keepspaces=true,showstringspaces=false]
SYSTEM EXCERPT
Given the instruction, measured scene subjects, selected motion
evidence, and the approved opening RGB/camera, return ONE
operation timeline.
Return JSON: {"operations":[{"op_id":"unique_id",
"method":"exact codebook name","start_sec":number,
"end_sec":number,"parameters":{}}]}.
Independent channels may overlap with independent time windows.
The approved opening pose and lens cannot change.
Validation errors and visual critiques are repair evidence:
return a complete corrected timeline.
\end{lstlisting}

The logged initial response $\mathcal{P}^{(0)}$ contains the following four concurrent operations. This display retains the principal parameters; all four use smoothstep easing, and subject-referenced operations target \texttt{hero\_01} at its face anchor.

\begin{lstlisting}[basicstyle=\ttfamily\footnotesize,breaklines=true,breakatwhitespace=false,columns=fullflexible,keepspaces=true,showstringspaces=false]
INITIAL PLAN P^(0)
Screen Position Lock  [0, 9] s   screen_uv=(0.50, 0.56)
                                 transition_sec=1.0
                                 smoothing_sec=0
Zoom                  [0, 9] s   focal_target=43.3747 mm
Orbit                 [0, 9] s   left, angle=120 degrees
                                 radius_scale=1.0
Dolly In              [0, 9] s   distance_scale=0.72
\end{lstlisting}

The executor composes these operations into camera states and renders $R(S,\mathcal{P}^{(0)})$. The initial plan attempts to combine forward movement and lens widening with a large orbit; whether this combination realizes the instruction is assessed from its execution and rendered result.

\paragraph{Motion Critic and Director revision}

The motion Critic receives the instruction, operation timeline, execution measurements, and ordered rendered frames. These observations instantiate $R(S,\mathcal{P}^{(0)})$ in the method overview. Both
evaluations in this example use the full budget of 10 frames, drawn from uniform
samples and the start, midpoint, and end of each operation; frames at 5\,s and 10\,s
are shown below as representative samples.

\begin{lstlisting}[basicstyle=\ttfamily\footnotesize,breaklines=true,breakatwhitespace=false,columns=fullflexible,keepspaces=true,showstringspaces=false]
SYSTEM EXCERPT
Review one continuous camera trajectory from ordered RGB frames
and exact screen-space metrics.
Judge requested movement by rendered evidence and measured motion,
not literal operation names.
Be specific about the operation and frame where a problem occurs.
Return accept, scores, failure_tags, diagnosis,
operation_feedback, and repair_patches.
\end{lstlisting}

The initial evaluation produces the following six scores. The equally weighted average gives $q_{\mathrm{motion}}^{(0)}=0.5883$, below the 0.62 acceptance threshold, even though the model's Boolean \texttt{accept} field is true.

\begin{lstlisting}[basicstyle=\ttfamily\footnotesize,breaklines=true,breakatwhitespace=false,columns=fullflexible,keepspaces=true,showstringspaces=false]
{
  "accept": true,
  "scores": {
    "prompt_alignment": 0.43,
    "subject_readability": 0.78,
    "composition": 0.57,
    "spatial_storytelling": 0.48,
    "motion_coherence": 0.72,
    "cinematic_quality": 0.55
  }
}
\end{lstlisting}

The following excerpt from the logged diagnosis forms part of $\mathcal{F}_{\mathrm{motion}}^{(0)}$:

\begin{lstlisting}[basicstyle=\ttfamily\footnotesize,breaklines=true,breakatwhitespace=false,columns=fullflexible,keepspaces=true,showstringspaces=false]
The figure remains readable with the entire hairstyle visible and approximately stable head size, but the shot reads primarily as a large orbit rather than a forward dolly-zoom. The divider begins on the right, passes behind the torso, then becomes a nearly horizontal barrier instead of a left-side receding depth cue. The ending settles into a rear three-quarter view with distracting cars and another figure. The camera appears to settle, but releasing the framing lock allows the subject to drift during the requested final portrait hold.
\end{lstlisting}

The Director's refinement request retains $\mathcal{G}$, $\mathcal{B}$, and $c^*$ and adds the previous timeline and feedback. The payload uses \texttt{previous\_timeline} for $\mathcal{P}^{(0)}$ and \texttt{feedback} for the evaluation and execution diagnoses; relevant rendered frames accompany the request. The instruction to return a complete corrected timeline yields $\mathcal{P}^{(1)}$:

\begin{lstlisting}[basicstyle=\ttfamily\footnotesize,breaklines=true,breakatwhitespace=false,columns=fullflexible,keepspaces=true,showstringspaces=false]
REVISED PLAN P^(1)
Screen Position Lock  [0, 10] s  screen_uv=(0.50, 0.56)
                                 transition_sec=1.0
                                 smoothing_sec=0
Dolly In              [0, 9] s   distance_scale=0.72
Zoom                  [0, 9] s   focal_target=43.3747 mm
Orbit                 [0, 2.5] s left, angle=18 degrees
                                 radius_scale=1.0
\end{lstlisting}

The revised plan reduces the orbit from $120^\circ$ over nine seconds to $18^\circ$ over the first 2.5 seconds, while extending the framing lock through the full shot. The opening is unchanged. Re-evaluation gives $q_{\mathrm{motion}}^{(1)}=0.7367$, and the run selects this plan. The Critic still records divider overlap and residual drift during the final hold; acceptance does not imply that every requested compositional detail has been satisfied.

\begin{lstlisting}[basicstyle=\ttfamily\footnotesize,breaklines=true,breakatwhitespace=false,columns=fullflexible,keepspaces=true,showstringspaces=false]
{
  "accept": true,
  "scores": {
    "prompt_alignment": 0.67,
    "subject_readability": 0.88,
    "composition": 0.65,
    "spatial_storytelling": 0.72,
    "motion_coherence": 0.8,
    "cinematic_quality": 0.7
  }
}
\end{lstlisting}

\paragraph{Extension to video-driven generation}

For V2V, $x$ is a reference video. Reference analysis extracts camera-motion and framing cues, which the Interpreter grounds in the target scene. The resulting script $\mathcal{B}$ is augmented with reference evidence for the relevant agents. The Scout uses reference imagery to guide opening composition, with LingBot-derived relative poses providing auxiliary motion context. The Director retains access to timestamped reference frames, reference analysis, and relative-pose evidence throughout planning and refinement. Because the reconstructed translation has unverified metric scale, these poses provide motion cues rather than target-scene camera coordinates. The motion Critic directly compares timestamped reference and rendered frames, assessing operation, direction, amplitude, timing, and endpoint fidelity, together with motion smoothness; it does not receive LingBot poses. Both settings share the atomic-operation representation and executor, with the selected opening fixed during motion-plan refinement, while using task-specific evaluation criteria.

\subsection{Atomic Camera Operations and Parameters}
\label{sec:supp_operations}

Table~\ref{tab:atomic_parameters} lists the 22 operations and their configurable parameters. Values are codebook defaults, which the Director may override when constructing a plan. Each operation also specifies its start and end times in seconds. Angles use degrees, focal lengths use millimeters, and explicit distances use scene world units. Dolly scales are endpoint-to-initial distance ratios; Truck and Pedestal scales multiply the scene-derived subject scale.

Common defaults are \texttt{easing=smoothstep} and, where exposed, \texttt{transition\_sec=0.5} and \texttt{smoothing\_sec=0}. Operations marked $\dagger$ additionally require a \texttt{target} and accept an \texttt{anchor}, defaulting to \texttt{center}, except Height Follow, Aim Follow, and Frame Subject, which default to \texttt{face}. Both focus operations use \texttt{aperture\_fstop=4.0} by default.

{\footnotesize
\setlength{\tabcolsep}{4pt}
\renewcommand{\arraystretch}{1.12}
\begin{longtable}{@{}>{\raggedright\arraybackslash}p{0.20\linewidth}>{\raggedright\arraybackslash}p{0.26\linewidth}>{\raggedright\arraybackslash}p{\dimexpr0.54\linewidth-16pt\relax}@{}}
\caption{Atomic operations, their effects, and operation-specific parameters. Common parameters are given above.}\label{tab:atomic_parameters}\\
\toprule
\textbf{Operation} & \textbf{Effect} & \textbf{Parameters (defaults)} \\
\midrule
\endfirsthead
\toprule
\textbf{Operation} & \textbf{Effect} & \textbf{Parameters (defaults)} \\
\midrule
\endhead
\bottomrule
\endfoot
Dolly In$\dagger$ & Move toward the starting target anchor. & \texttt{distance\_scale=0.72} ($0<s<1$). \\
Dolly Out$\dagger$ & Move away from the starting target anchor. & \texttt{distance\_scale=1.35} ($s>1$). \\
Truck & Translate along the initial camera-right axis. & \texttt{distance\_scale=0.75}; \texttt{direction=right}. \\
Pedestal Up & Translate along world $+Z$. & \texttt{height\_scale=0.65}. \\
Pedestal Down & Translate along world $-Z$. & \texttt{height\_scale=-0.55}. \\
Orbit$\dagger$ & Circle a target in the world $XY$ plane. & \texttt{orbit\_angle\_deg=28}; \texttt{direction=right}; \texttt{radius\_scale=1.0}. \\
Path Follow & Traverse an existing scene curve. & \texttt{path\_id} required; \texttt{progress\_start=0}; \texttt{progress\_end=1}; \texttt{speed\_mode=constant\_arc\_length}. \\
Position Follow$\dagger$ & Maintain a target-relative positional offset. & \texttt{axes=[x,y,z]}; \texttt{offset\_mode=initial}; \texttt{offset=[0,0,0]} for specified mode; \texttt{offset\_space=world}. \\
Height Follow$\dagger$ & Track target-anchor height plus a target-height-scaled offset. & \texttt{height\_offset\_scales=0.02}. \\
Distance Follow$\dagger$ & Maintain camera-to-target distance by radial correction. & \texttt{distance\_mode=initial}; \texttt{distance} required for specified mode; \texttt{composition\_policy=exclusive}. \\
Pan & Rotate aim about world up. & \texttt{angle\_deg=20}. \\
Tilt & Rotate aim about camera right. & \texttt{angle\_deg=15}. \\
Aim Follow$\dagger$ & Direct the viewing axis toward a target anchor. & Target, anchor, and common transition parameter. \\
Screen Position Lock$\dagger$ & Keep a target anchor at a normalized image location. & \texttt{screen\_uv=[0.5,0.5]}; image origin is bottom-left. \\
Roll & Rotate about the optical axis. & \texttt{angle\_deg=15}; signed, unwrapped angle. \\
Horizon Lock & Level the image relative to world up. & \texttt{world\_up=[0,0,1]}. \\
Zoom & Interpolate focal length in log space. & \texttt{focal\_target=50}. \\
Frame Subject$\dagger$ & Adjust focal length to maintain a desired subject image scale. & \texttt{portrait\_height\_fraction=0.65}; \texttt{portrait\_screen\_fraction=0.78}. \\
Focus Pull & Transfer focus between two subject anchors. & \texttt{from\_target}, \texttt{to\_target} required; \texttt{from\_anchor=center}; \texttt{to\_anchor=center}; \texttt{track\_endpoints=true}; \texttt{composition\_policy=exclusive}. \\
Focus Follow$\dagger$ & Track a target's positive camera-space focus depth. & Target, anchor, and common transition, smoothing, and aperture parameters. \\
Shake / Handheld & Add seeded position and angular perturbations. & \texttt{seed=42}; \texttt{translation\_amplitude=[0.01,0.01,0.01]}; \texttt{rotation\_amplitude\_deg=[0.3,0.3,0]}; \texttt{frequency\_hz=1.5}; \texttt{space=camera}; \texttt{fade\_in\_fraction=0.1}; \texttt{fade\_out\_fraction=0.1}. \\
Static Shot & Hold the incoming camera state. & No operation-specific parameters. \\
\end{longtable}
}

\paragraph{Definitions of $\ominus$ and $\oplus$}
The operator $\ominus$ extracts a typed relative change from an operation's absolute track $\hat c_i(t)$ and its reference state at $t_i^{\mathrm{ref}}$, which is the operation start or the applicable composition boundary. Translational offsets use vector differences, angular offsets use signed angle differences about their declared axes with full turns retained, and positive scale quantities use ratios:
\begin{equation}
\Delta p_i=\hat p_i(t)-\hat p_i(t_i^{\mathrm{ref}}),\qquad
\Delta\alpha_i=\hat\alpha_i(t)-\hat\alpha_i(t_i^{\mathrm{ref}}),\qquad
r_{\gamma,i}=\frac{\hat \gamma_i(t)}{\hat \gamma_i(t_i^{\mathrm{ref}})}.
\end{equation}
Here $\gamma$ denotes radial distance, orbit radius, focal length, or focus depth, as applicable. Orbit combines a horizontal angular offset with a radius ratio; it is not represented as a simple additive position offset.

The operator $\oplus$ applies compatible relative changes to the reference or tracking-derived camera state. Translation offsets add, compatible angular offsets add within their prescribed axes, and positive scale ratios multiply. Orbit and Dolly transforms precede additive Truck and Pedestal offsets; aim composition applies base aim, Pan, Tilt, and then roll when constructing the orientation. Zoom ratios multiply the focal-length base, and explicitly enabled focus-offset ratios multiply the focus base. These are channel-specific composition rules, rather than elementwise addition of complete camera states.

\subsection{Hyperparameters and Computational Cost}
\label{sec:supp_cost}
All agents use \texttt{gpt-6-astra} with medium reasoning effort. The Scout evaluates up to six rendered candidates per round for at most two rounds. The Director allows three additional Critic-guided refinement rounds and up to two validation-driven repairs per planning invocation. The opening-view and motion acceptance thresholds are set to $0.70$ and $0.62$, respectively. The motion Critic observes at most 10 rendered frames per evaluation in T2V and at
   most 16 reference plus 16 rendered frames in V2V; the opening Critic observes one
   rendered image per candidate. FilmFlow requires approximately 2.23 minutes and 69.5k tokens per task on average, comprising 65.9k input and 3.7k output tokens across 6.23 model calls. These averages include unsuccessful attempts and cover scene grounding, view selection, motion planning, and validation, excluding final video export.

\section{CinemaBench}
\label{sec:supp_benchmark}

\subsection{Dataset Composition}
\label{sec:supp_benchmark_composition}

\textbf{CinemaBench.} CinemaBench contains 150 Blender scenes built from self-created and licensed assets,
including 73 dynamic scenes with animated subjects and 77 static scenes.
The scenes span diverse indoor and outdoor environments with single or multiple subjects,
providing varied spatial layouts, occlusion patterns, and subject motion for camera planning.
Source attribution and license information for all assets will be included in the release.

\paragraph{Text-driven tasks (T2V).}
Each scene is paired with five natural-language camera instructions, yielding 750 tasks.
Each T2V task consists of a target scene and an instruction; the target scene supplies the
subjects, geometry, and animation on which the generated trajectory is executed.
Per scene, the five instructions comprise one easy, two medium, and two hard tasks.
Easy tasks involve a single subject and a single movement; medium tasks combine two or more
movements or subjects; hard tasks require coordinated concurrent operations, attention shifts
across multiple subjects, or precise temporal constraints.

The authors manually wrote 50 instructions.
The remaining 700 were generated with \texttt{gpt-6-astra}, prompted with the human-written
examples and conditioned on scene geometry, available objects, and existing animation.
Each generated instruction passes programmatic checks, which verify that referenced objects
exist, that timing is consistent with the shot duration, and that the difficulty label holds.
It then undergoes a model-based review for consistency with scene evidence.
Instructions that failed either stage were regenerated until they passed.

\paragraph{Reference-video tasks (V2V).}
A subset of 60 scenes is additionally paired with 60 reference clips.
Each V2V task consists of a target scene and a reference clip whose camera behavior serves as
the generation condition.
The clips are drawn from excerpts of publicly released films and self-recorded footage,
and last 10 seconds on average.
Reference and generated videos may differ in scene appearance; evaluation concerns the
reference camera behavior and its realization in the target scene.

\paragraph{Split.}
We split CinemaBench by scene into 50 development scenes (250 T2V and 20 V2V tasks) and
100 test scenes (500 T2V and 40 V2V tasks), with no scene shared across the two splits.
Development scenes are used only to build the ICL example bank (Sec.~\ref{sec:icl});
all reported results use the test split.
Table~\ref{tab:bench_stats} summarizes the statistics per split.

\begin{table}[h]
\centering
\caption{CinemaBench statistics.}
\label{tab:bench_stats}
\small
\begin{tabular}{lccc}
\toprule
 & Dev & Test & Total \\
\midrule
Scenes (dynamic / static) & 27 / 23 & 46 / 54 & 73 / 77 \\
T2V tasks (easy / medium / hard) & 50 / 100 / 100 & 100 / 200 / 200 & 150 / 300 / 300 \\
V2V tasks & 20 & 40 & 60 \\
Human-written instructions & 20 & 30 & 50 \\
\bottomrule
\end{tabular}
\end{table}

\subsection{Language of Camera Motion}
\label{sec:supp_camera_language}

CinemaBench instructions describe camera behavior along four aspects: subjects, motion,
framing, and temporal structure.
\emph{Subject} descriptions specify what the camera should observe and the spatial references
around which it moves.
\emph{Motion} descriptions specify actions such as approaching, retreating, orbiting, rising,
panning, or following, together with their direction and extent where needed.
\emph{Framing} descriptions specify subject placement, scale, visibility, or a change of
attention.

\emph{Temporal} expressions distinguish sequential from concurrent actions.
``Orbit the subject, then move closer'' specifies an ordered transition, whereas ``orbit the
subject while moving closer'' requires rotation around the subject and radial approach at the
same time.
Instructions may also specify partial overlap, such as beginning a push-in before an orbit ends.
A valid trajectory must realize both the constituent actions and their temporal organization.
This is the capability tested by the Sequential-only ablation (Sec.~\ref{sec:ablation}).

Instructions also express visual intent beyond named motions.
Requests to reveal a subject, shift attention, or hold a close-up describe the desired visual
outcome without prescribing a unique operation sequence.
Several combinations of atomic operations can therefore satisfy the same instruction.
CinemaBench accordingly specifies no ground-truth operation sequence or trajectory for T2V;
fulfillment is judged from the executed camera behavior and the rendered composition.

\subsection{Baseline Implementation}
All baselines receive the same inputs as FilmFlow. None is initialized from FilmFlow's
outputs. GenDoP and LAMP
produce trajectories in their own relative coordinate frames; we anchor frame~0 to the
full pose and focal length of the scene camera at the task's start frame, rotate the
native frame-0 camera basis to match its orientation, and apply subsequent relative
displacements under the same rotation, preserving roll. Trajectory scale is derived from
the task's target objects: we average the largest world-space bounding-box edge of each
target at the start frame, clamped below at 0.3, to obtain a subject scale $s$; GenDoP
trajectories are scaled so that their longest axis spans $2s$, and LAMP trajectories by
$s/64$, where 64 is LAMP's native coordinate unit. Native focal lengths, when provided,
are converted assuming a 36\,mm sensor; otherwise the scene camera's lens is used.
CinemaTraj plans its own opening camera from the scene, as in its original design. For V2V, LingBot-Map and VGGT-$\Omega$ serve as direct pose-transfer baselines: their
estimated camera trajectories are converted from native axis conventions to Blender's
and applied in the target scene without scene-aware placement, VLM-based motion
planning, or closed-loop refinement. The zero-shot VLM baseline
uses gpt-6-astra with the same reference frames as FilmFlow, directly outputs camera
keyframes, and interpolates between them to the full shot. GenDoP and LAMP do not track
moving subjects, which reflects a limitation of these methods rather than of the
adaptation.

\section{Additional Qualitative Results}
\label{sec:supp_additional_results}

\begin{figure*}[h!]
\centering
\includegraphics[width=\textwidth]{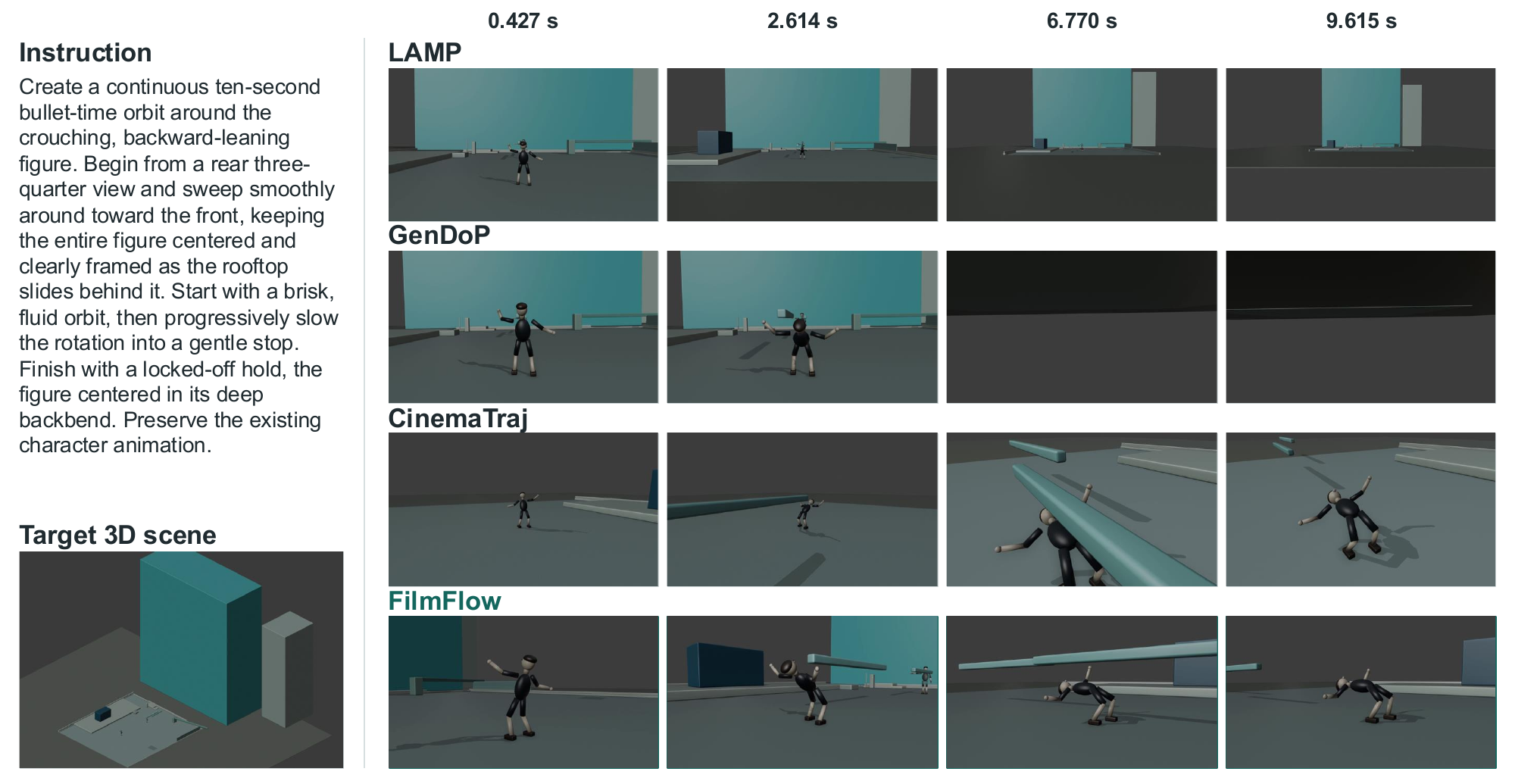}
\caption{
Additional qualitative comparison for text-directed camera generation.
The instruction requests a bullet-time orbit that starts from a rear three-quarter view,
sweeps to the front of a crouching, backward-leaning figure, and settles into a locked-off hold.
The instruction and target 3D scene are shown on the left; rows show each method's output at
the indicated timestamps. LAMP and GenDoP drift away from the figure toward the wall or empty
frames, and CinemaTraj approaches the figure but leaves it off-center under an oblique view.
FilmFlow keeps the figure centered as the rooftop slides behind it and ends on a stable frontal
view of the backbend.
}
\label{fig:supp_t2v_comparison}
\end{figure*}

\begin{figure*}[h!]
\centering
\includegraphics[width=\textwidth]{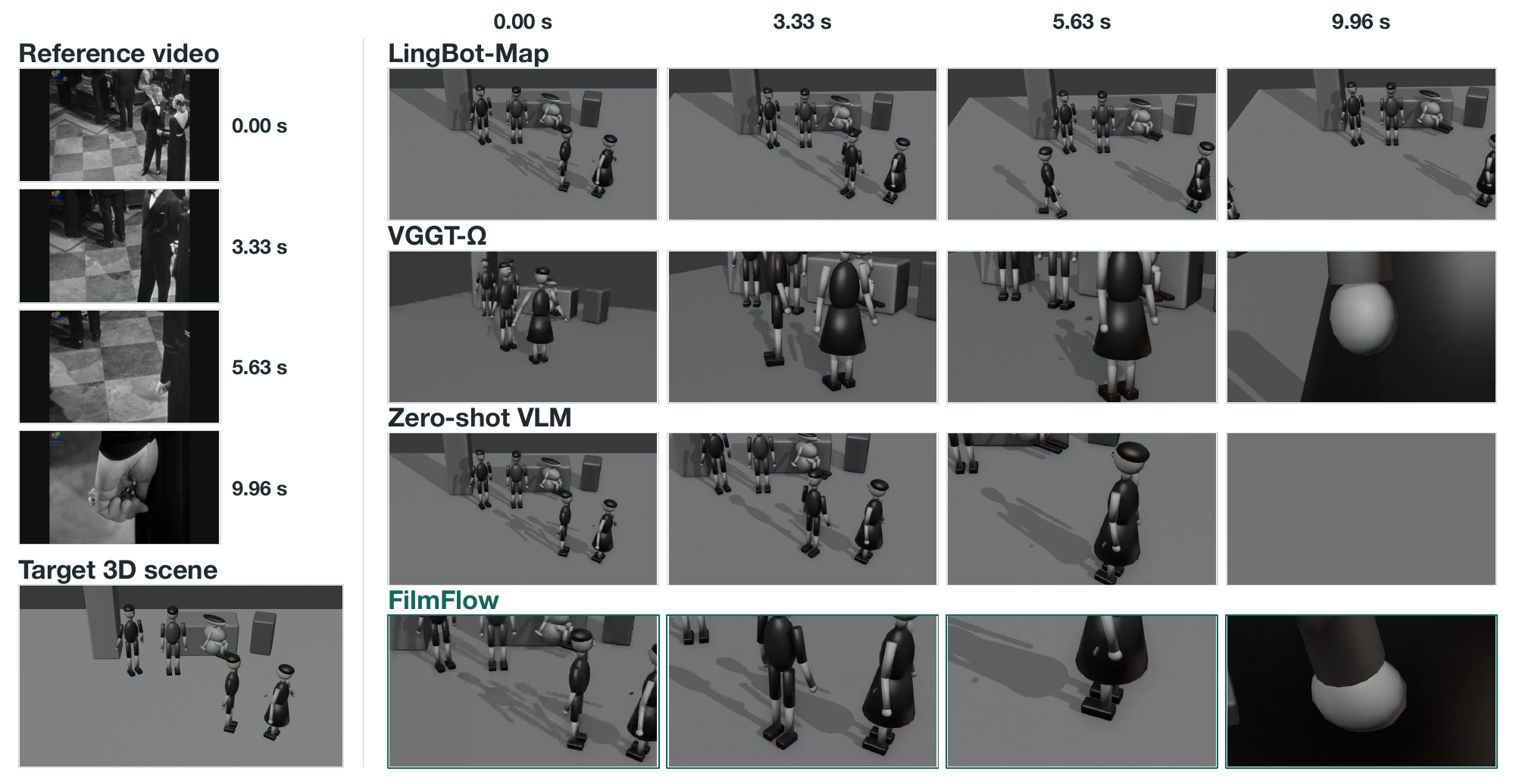}

\caption{Additional qualitative comparison for video reference-directed camera generation.
The reference clip moves from a view of the group toward a single subject and ends on a tight
close-up. Reference frames and the target 3D scene are shown on the left; rows show each
method's output at the indicated timestamps. LingBot-Map stays close to a static wide view,
VGGT-$\Omega$ crops the subject in intermediate frames, and the zero-shot VLM loses the subject
in the final frame. FilmFlow reproduces the progression from group view to single subject and
ends on a close-up that matches the reference framing. The comparison concerns camera behavior
rather than scene appearance.}
\label{fig:supp_v2v_comparison}
\end{figure*}

Figures~\ref{fig:supp_t2v_comparison}
and~\ref{fig:supp_v2v_comparison} provide additional comparisons
for text-directed and video reference-directed camera generation.
The examples complement those in the main paper with different
scenes and camera-motion requirements.
Within each example, all methods receive the same target scene
and conditioning input.
The sampled frames illustrate the progression of camera motion,
subject framing, and composition, while the corresponding
supplementary videos provide the complete temporal context.


\section{Case Study: Scout Refinement}
\label{sec:supp_scout_case}

\begin{figure*}[h!]
\centering
\includegraphics[width=\textwidth]{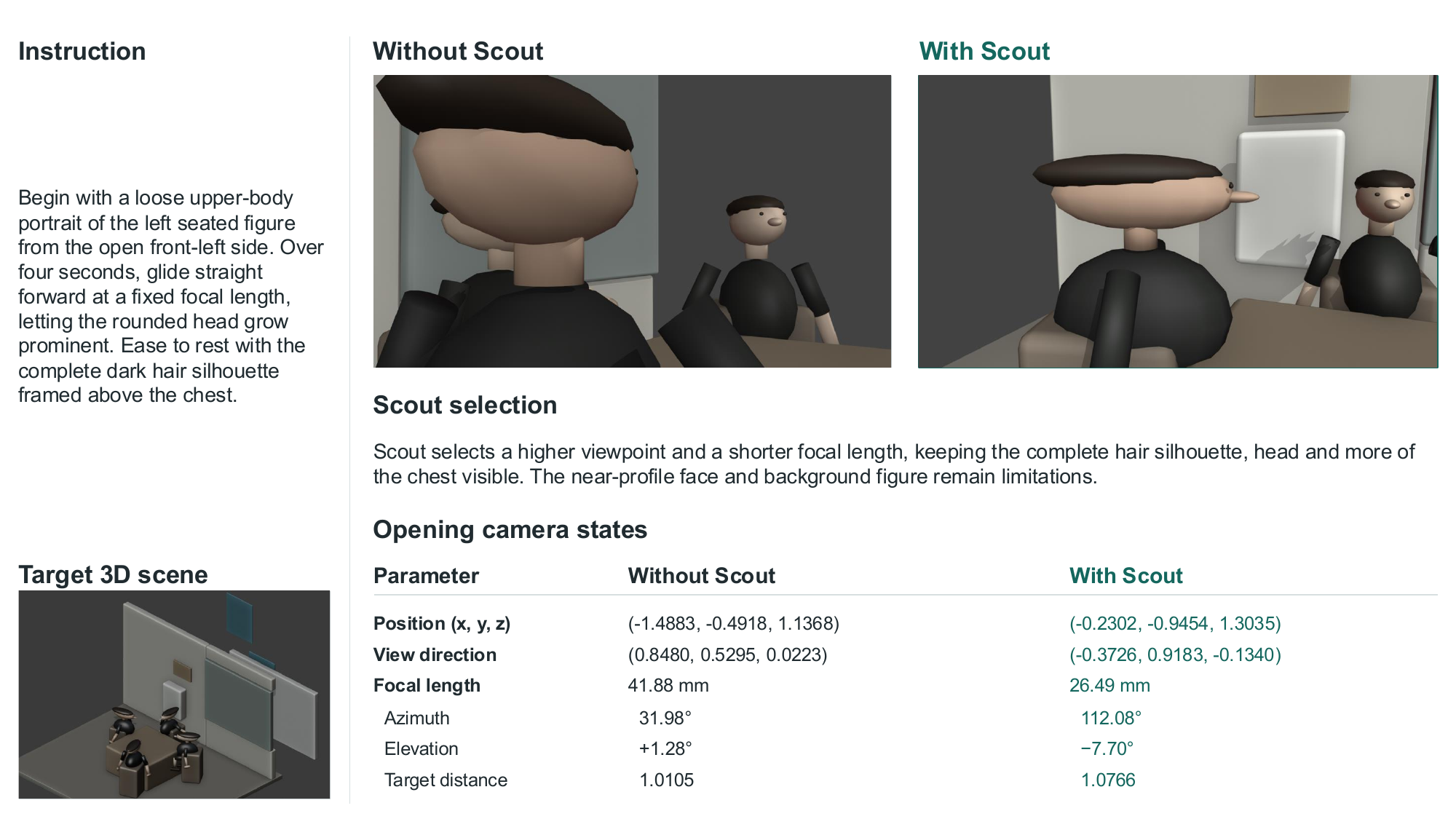}
\caption{
An example of Scout refinement. The Director-specified opening (w/o Scout) and
the Scout-selected opening are shown for the same instruction, with the
Scout's selection rationale and the corresponding camera states.
Elevation denotes the camera's pitch angle; the Scout-selected camera
is placed higher and pitched downward toward the subject.
}
\label{fig:supp_scout_refinement}
\end{figure*}

Figure~\ref{fig:supp_scout_refinement} compares the opening specified directly by the Director (w/o Scout) with the
Scout-selected opening for an instruction requesting a loose upper-body portrait that
glides forward at a fixed focal length. The Director-specified 41.88\,mm opening crops
the top of the hair and shows little of the chest; since the focal length is fixed, the
forward glide can only tighten this framing further. The Scout
instead selects a higher viewpoint pitched slightly downward (elevation
$-7.70^\circ$) with a shorter 26.49\,mm lens, revealing the complete
hair silhouette, head, and more of the chest, and leaving room for the
requested push-in. This illustrates why the opening is selected before
motion planning: once fixed, a poor initial state constrains every
subsequent operation.

\section{Case Study: Director Refinement}
\label{sec:supp_director_case}

\begin{figure*}[h!]
\centering
\includegraphics[width=\textwidth]{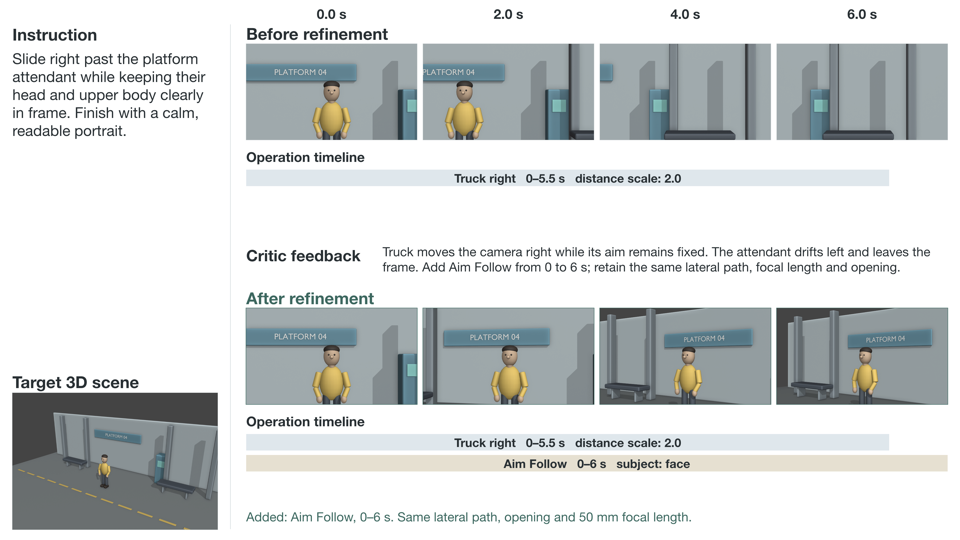}
\caption{
An example of Director refinement under a fixed opening.
Frame sequences and operation timelines before and after revision
show how Critic feedback guides changes to operation parameters
and temporal composition.
}
\label{fig:supp_director_refinement}
\end{figure*}

Figure~\ref{fig:supp_director_refinement} examines operation-plan
refinement after opening-view selection.
The initial and revised plans share the same instruction,
target scene, and fixed opening.
Frames are displayed at corresponding timestamps, and the
associated operation timelines show the active intervals
of the constituent motions.

The Critic feedback links the observed execution problem to
specific operations.
The before-and-after timelines identify changes to operation
selection, parameters, or temporal intervals.
Displaying the rendered frames together with these changes
allows the revision to be assessed in terms of both its
visual effect and its underlying operation composition.
The accompanying videos provide the temporal evidence needed
to inspect motion continuity between the sampled frames.

\section{Case Study: Generalization through ICL}
\label{sec:supp_icl_case}

\begin{figure*}[h!]
\centering
\includegraphics[width=\textwidth]{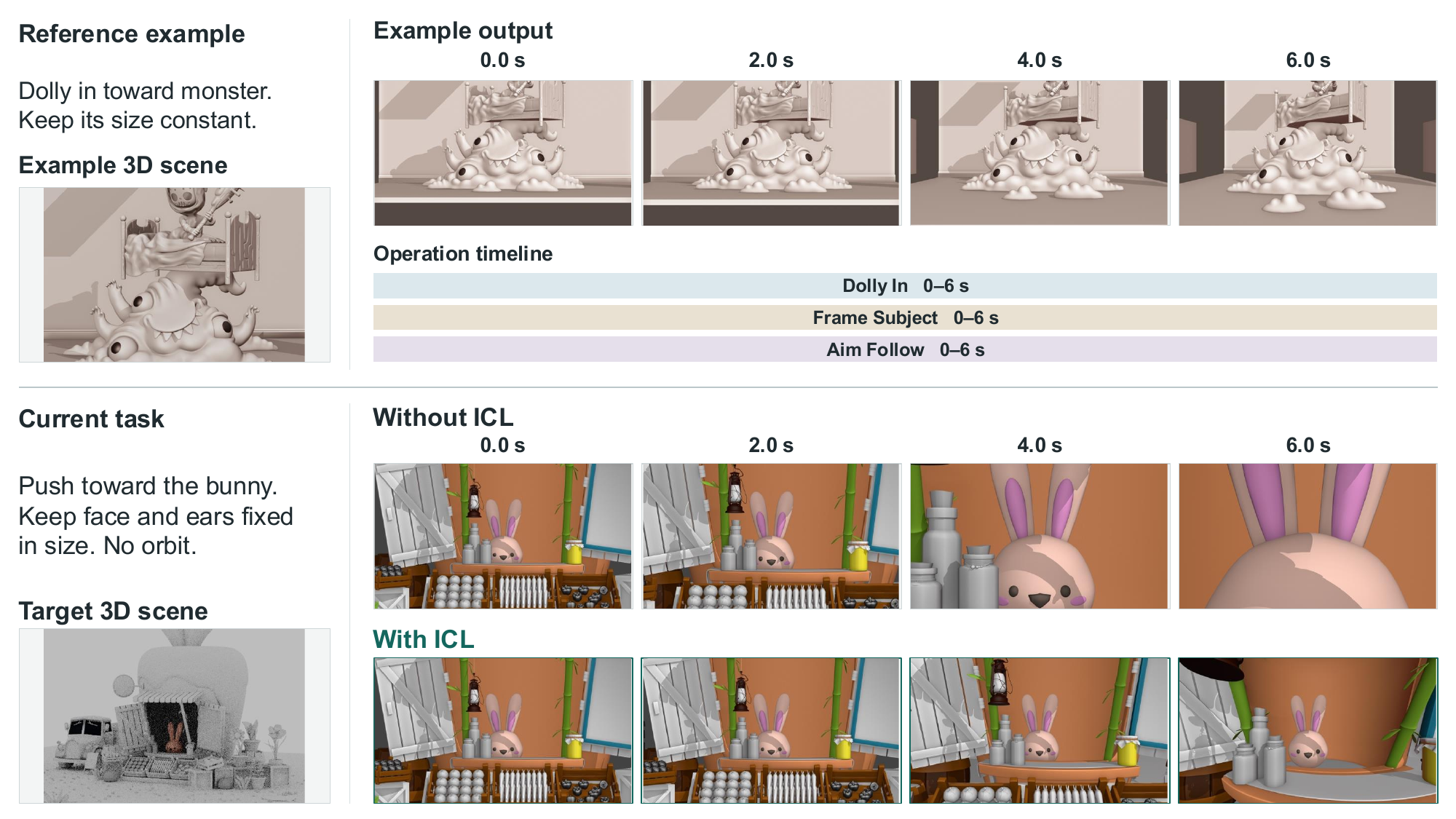}
\caption{
An example of in-context conditioning across different scenes.
The retrieved demonstration is displayed separately from the
current task. Results with and without ICL use the same current
instruction and target scene.
}
\label{fig:supp_icl_generalization}
\end{figure*}

Figure~\ref{fig:supp_icl_generalization} shows how a retrieved
demonstration guides planning in a different scene.
The retrieved example asks the camera to dolly in toward a monster
while keeping its on-screen size constant. Its validated plan
realizes this constraint by running three concurrent operations
over the full shot: Dolly In moves the camera forward, Frame Subject
compensates the focal length to hold the subject's framed size,
and Aim Follow keeps the subject centered.
The current task, ``Push toward the bunny. Keep face and ears fixed
in size. No orbit,'' shares this intent but differs in scene layout,
subject, and scale. The example is retrieved through overlapping
intent keywords rather than object identities.

Without ICL, the Director plans a forward push alone. The bunny grows
steadily in frame and fills the view by 6.0\,s, violating the
size constraint.
With ICL, the Director adopts the demonstrated composition,
pairing the forward motion with Frame Subject and Aim Follow,
so the bunny keeps a nearly constant framed size while nearby props
grow as the camera advances, producing a dolly-zoom effect.
The demonstration transfers only the operation choice and temporal
structure; target references, distances, and focal lengths are
resolved in the current scene.
This case study illustrates the conditioning mechanism and
complements the quantitative ICL ablation in Table~\ref{tab:ablation}.

\section{Application: Grey Model Reference as Video-Generation Conditions}
\label{sec:supp_video_gen}

\begin{figure}[t]
\centering
\includegraphics[width=\linewidth]{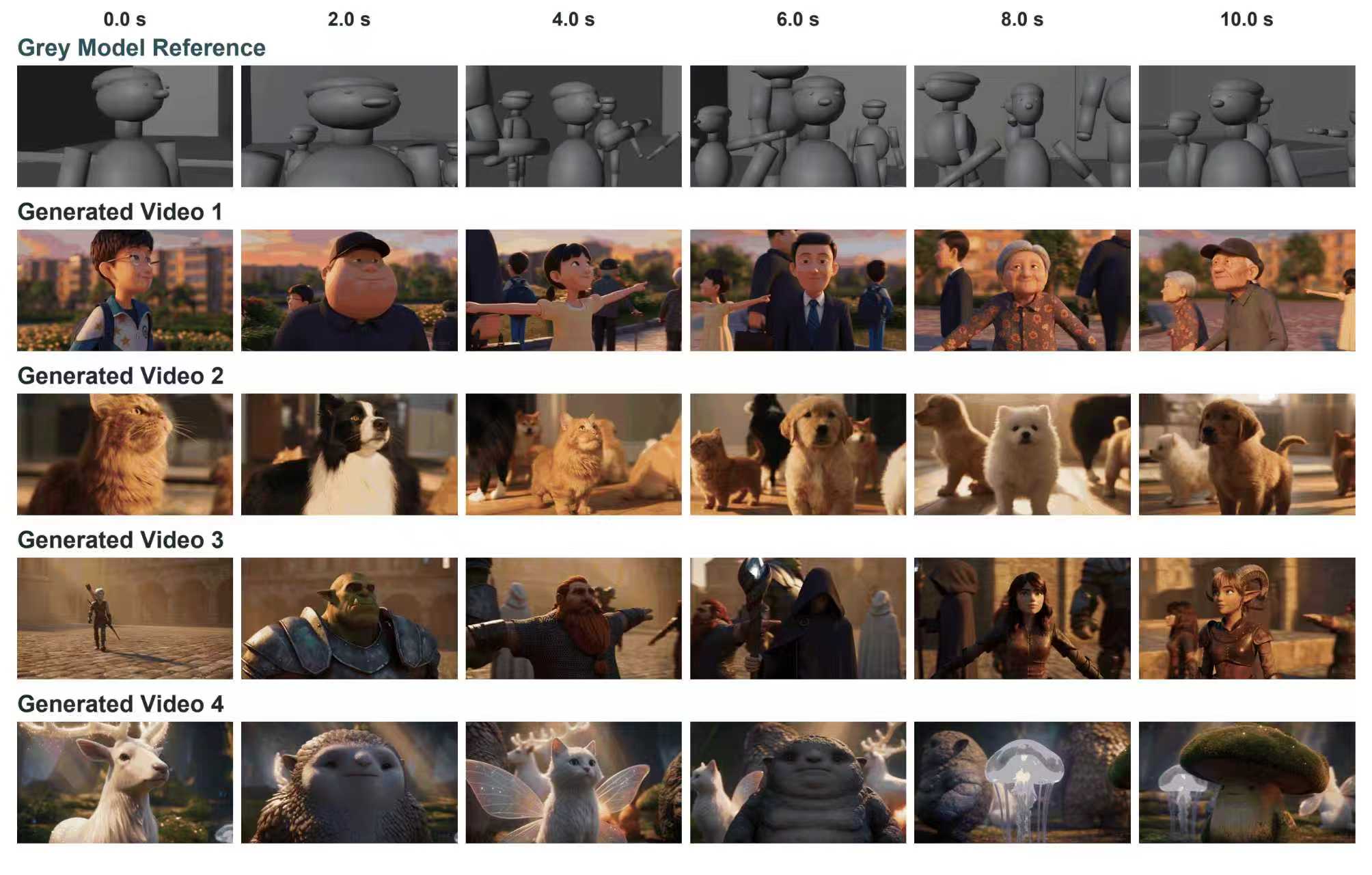}
\caption{\textbf{Grey model render as a video-generation condition (group close-ups).}
The top row shows the grey model render of a FilmFlow trajectory that gives successive
close-ups to figures in a group. Each subsequent row is generated from the same render with
a different text prompt. Camera motion, framing, and subject placement follow the render,
while appearance follows the prompt.}
\label{fig:clay_gen_orbit}
\end{figure}

\begin{figure}[t]
\centering
\includegraphics[width=\linewidth]{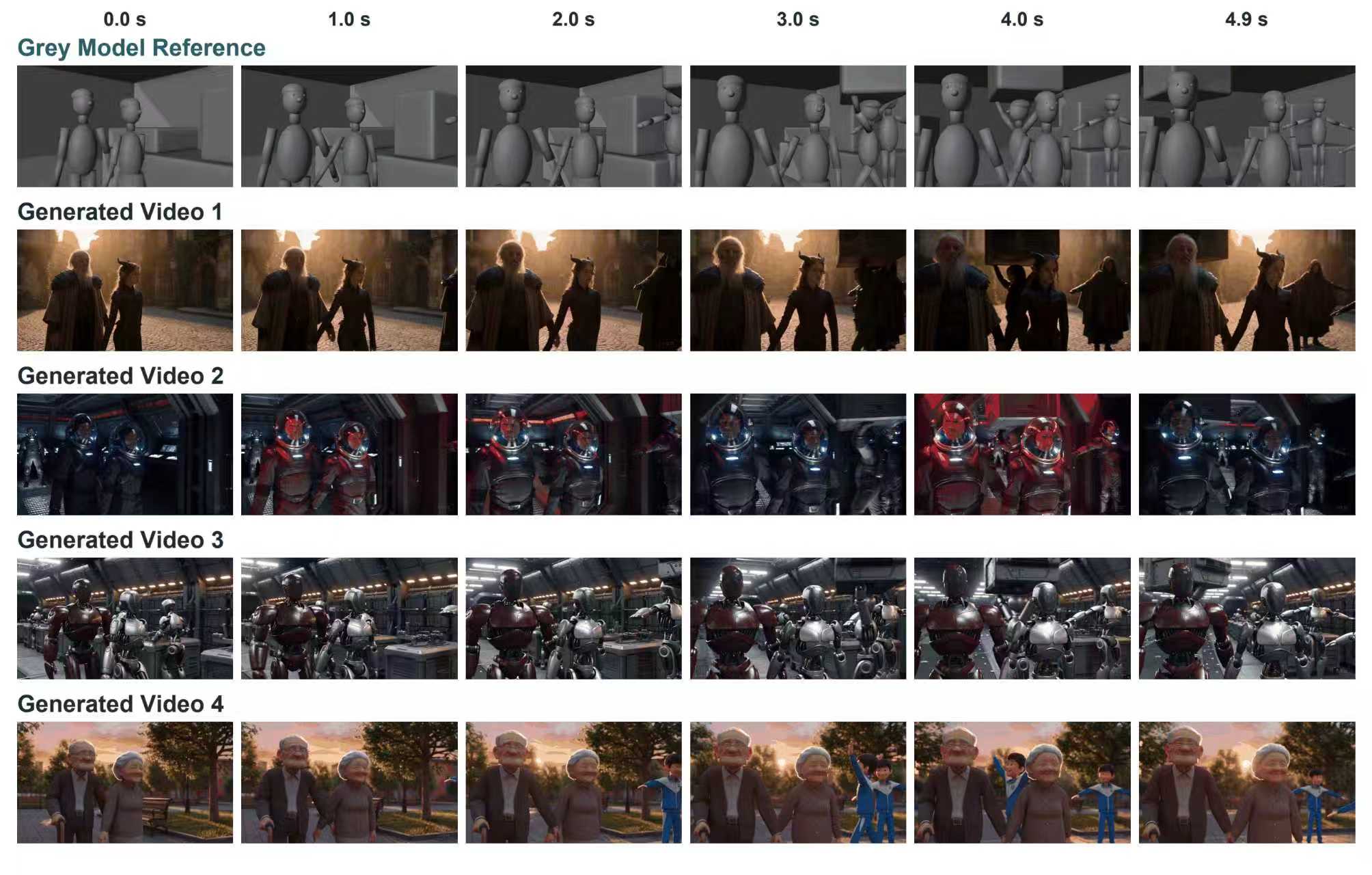}
\caption{\textbf{Grey model render as a video-generation condition (leading shot).}
The top row shows the grey model render of a FilmFlow trajectory that leads two walking
subjects while other characters enter from the right. The four generated videos keep the
framing of the pair and the timing of the entering characters across different styles.}
\label{fig:clay_gen_track}
\end{figure}

Beyond executable trajectories, FilmFlow's grey model renders can serve as structural
conditions for video generation. Because a render encodes camera motion, scene layout, and
subject animation without surface appearance, one camera plan can be reused across visual
styles by varying only the text prompt. Figures~\ref{fig:clay_gen_orbit}
and~\ref{fig:clay_gen_track} show two renders, each paired with four videos generated by
Seedance~\citep{seedance2026seedance}. In the first example, the camera gives successive close-ups
to figures in a group; in the second, it leads two walking subjects while other characters
enter from the right. Across stylized humans, animals, robots, and fantasy creatures, the
generated videos preserve the planned framing, subject placement, and timing, while their
appearance follows the prompt. This decoupling suggests that FilmFlow can act as a camera
planner for downstream generative filmmaking.

\end{document}